\documentclass[11pt]{article}
\usepackage[utf8]{inputenc}
\usepackage[margin=1in]{geometry}
\usepackage{amsmath}
\usepackage{amssymb}
\usepackage{xcolor}
\usepackage{makecell,float} 
\usepackage{mathtools}
\usepackage{authblk}
\usepackage{hyperref}
\usepackage{caption}
\usepackage{algorithm}
\usepackage{algpseudocode}
\usepackage{subcaption}
\usepackage{enumerate}
\usepackage{multirow}
\usepackage{adjustbox}
\usepackage{comment}
\usepackage{makecell}
\usepackage{amsthm}
\theoremstyle{remark}
\newtheorem{remark}{Remark}
\usepackage{stmaryrd}

\usepackage[round]{natbib}
\usepackage{xcolor}
\usepackage{authblk}
\usepackage{epstopdf}
\AppendGraphicsExtensions{.tiff}

\title{\vspace{-3ex}Tensorized Multi-Task Learning for Personalized Modeling of Heterogeneous Individuals with High-Dimensional Data}
\author[1]{Elif Konyar}
\author[2]{Mostafa Reisi Gahrooei}
\author[1]{Kamran Paynabar}
\affil[1]{Department of Industrial and Systems
Engineering, Georgia Institute of Technology, Atlanta, GA}
\affil[2]{Department of Industrial and Systems
Engineering, University of Florida, Gainesville, FL}
\date{\vspace{-5ex}}
\begin{document}

\maketitle

\begin{abstract}
Effective modeling of heterogeneous subpopulations presents a significant challenge due to variations in individual characteristics and behaviors. This paper proposes a novel approach to address this issue through multi-task learning (MTL) and low-rank tensor decomposition techniques. Our MTL approach aims to enhance personalized modeling by leveraging shared structures among similar tasks while accounting for distinct subpopulation-specific variations. We introduce a framework where low-rank decomposition decomposes the collection of task model parameters into a low-rank structure that captures commonalities and variations across tasks and subpopulations. This approach allows for efficient learning of personalized models by sharing knowledge between similar tasks while preserving the unique characteristics of each subpopulation. Experimental results in simulation and case study datasets demonstrate the superior performance of the proposed method compared to several benchmarks, particularly in scenarios with high variability among subpopulations. The proposed framework not only improves prediction accuracy but also enhances interpretability by revealing underlying patterns that contribute to the personalization of models.
\end{abstract}

\section{Introduction}\label{sec:Intro}

\sloppy
Model personalization, with broad applications in several fields, including precision medicine and healthcare \citep{johnson2021precision, hilton2020personalized, abul2019personalized}, advertising \citep{bilenko2011predictive}, and interactive user interfaces \citep{ma2021one}, involves tailoring models to account for the unique characteristics and features of individuals (or subgroups) within a population. A key challenge in achieving model personalization is addressing heterogeneity among individuals while leveraging their similarities. When each individual has access to a large amount of data, leveraging similarity is not essential, and one straightforward method is to fit separate models to each individual, allowing for fully individualized modeling. However, in most applications, including healthcare, access to a large sample size for each individual is difficult and expensive. An alternative approach to fully individualized modeling trains one model to fit all by combining data from all individuals. While this approach increases the sample size, it overlooks the unique traits of individuals and the variations between them. Therefore, middle-ground methods that can use shared information and capture data heterogeneity are necessary. 

An example of such a trade-off appears in telemonitoring applications for chronic disease management, such as remote assessment of Parkinson’s disease severity using smartphone-based sensor data. Telemonitoring enables the continuous and non-invasive monitoring of patients in their daily environments, providing valuable real-time insights into disease progression without the need for frequent clinical visits \citep{bot2016mpower}. However, effectively utilizing such data presents modeling challenges due to significant variability in symptoms among patients, where some may exhibit dominant tremors, while others primarily experience gait disturbances or motor slowness. Global models, which aggregate data across patients, often fail to capture these individualized patterns, resulting in less accurate predictions. At the same time, because patient-specific data is usually limited and noisy, fully individualized models cannot produce accurate predictions. Personalized models offer a practical and effective middle ground, which borrows strength from shared patterns in the population while tailoring predictions to each patient’s unique symptom profiles and disease trajectories. 

The challenge of limited data available to each individual is exacerbated when the data is high-dimensional. In many modern applications, data are collected in the form of structured high-dimensional data, including images or profiles that contain a large number of observations (features) per sample. The combination of high-dimensionality and limited data exacerbates the risk of overfitting and makes fully individualized modeling impractical. An example of such situations arises in neuroimaging-based diagnostic applications, such as identifying attention deficit hyperactivity disorder (ADHD) from functional magnetic resonance imaging (fMRI) data. In this setting, high-dimensional brain imaging data combined with highly limited samples per individual make it difficult to develop accurate diagnostic models. In addition, significant variability in brain activity patterns between individuals further complicates the modeling process. Personalized modeling of high-dimensional data can address these challenges by capturing individual-specific neural patterns while leveraging shared information across the population.

To mitigate the impact of limited data from heterogeneous individuals, traditional methods such as mixed-effect models \citep{oberg2007linear, verbeke1996linear, proust2005estimation} and the mixture of experts approach \citep{eavani2016capturing} have been introduced. Mixed-effect models address heterogeneity by incorporating random effects, which enables modeling individual variations. The mixture of experts utilizes multiple submodels to represent diverse subpopulations. However, a key assumption of these traditional approaches is that the variations in individual models are randomly distributed, which may not accurately reflect the complexities of real-world data. Furthermore, these methods cannot handle high-dimensional data structures. Given the limitations of mixed-effect and mixture of experts models, alternative methods such as transfer learning, collaborative learning, deep learning, and multi-task learning are gaining attention for model personalization. Transfer learning has been used to train personalized models based on previously trained models. These approaches often use models and data from other individuals to train a personalized model for a target individual \citep{zhang2021survey} and cannot simultaneously train personalized models for all individuals (or subgroups) of a population. Collaborative learning (CL) introduces a set of canonical models that represent various characteristics of heterogeneous populations \citep{lin2015domain, lin2017collaborative, feng2020learning}. Then, individual models are obtained as a convex combination of these canonical models. For instance, \cite{lin2017collaborative} propose a collaborative approach that assumes the existence of a group of canonical degradation models, with each model representing a different degradation mechanism. However, a significant challenge of these methods is determining the number of canonical models, which often requires domain knowledge that may not be available. In addition, the basis formed by these canonical models creates a constrained search space, which may not fully represent all possible models, especially when domain knowledge is limited. Finally, these models are mostly designed for linear and logistic regression frameworks and are not suitable for modeling structured high-dimensional data. Deep learning (DL) techniques offer a robust alternative for enhancing personalization by exploiting large and complex datasets \citep{gupta2023perspective}. These models are powerful in identifying patterns and relationships within data, which allows them to capture the unique characteristics of individuals or subpopulations. However, a significant drawback is their reliance on abundant data, which may not always be accessible, specifically in high-dimensional data settings. In addition, they are susceptible to ignoring the patterns of individuals with limited data. Additionally, deep learning models often lack interpretability, which is essential for practitioners in many fields, such as healthcare and agriculture.

Multi-task learning (MTL) has emerged as a framework that simultaneously learns a set of tasks (models) by leveraging the similarity between them \citep{zhang2021survey}. This concept is inspired by how humans learn, where knowledge gained in one area can assist in learning another, such as learning to drive a car can help in learning to drive a motorcycle. Motivated by this human ability, multi-task learning aims to enhance the generalization performance of multiple tasks by leveraging information shared between the tasks \citep{ zhou2015flexible, zhang2021survey}. The two main groups of MTL are feature-based and parameter-based methods. The feature-based techniques find a shared feature space invariant to all the tasks by applying deep neural networks \citep{caruana1997multitask,liao2005radial,li2014heterogeneous,liu2015multi,shinohara2016adversarial, liu2017adversarial} or other methods such as projection \citep{argyriou2007spectral, argyriou2008convex, titsias2011spike}.  The main limitation of these methods is that they are very sensitive to outlier tasks and require abundant data \citep{zhang2021survey}. The parameter-based methods have three main categories: Cluster-based, relationship-based, and subspace-based methods. The cluster-based ones use clustering algorithms to group the tasks into similar ones \citep{thrun1996discovering, crammer2012learning}. The main limitation of these methods is that they cannot capture the between-cluster structures. The relationship-based ones quantify the relationship between the tasks primarily by task similarity, correlation, or covariance \citep{evgeniou2004regularized, evgeniou2005learning, bonilla2007multi, zhang2012convex}. While these methods learn model parameters and pairwise task relations, they are computationally costly. Existing subspace-based methods learn a shared linear subspace between the task parameters \citep{ando2005framework, chen2009convex, han2016multi}. While powerful, these methods can only capture the linear relationship among tasks and features and cannot capture cross-correlations between subgroups and features, nor the hierarchical structure of the features. Furthermore, none of these methods can capture the complex correlation structure that may exist in high-dimensional data.

Multi-task learning (MTL) offers a promising framework for personalization by treating each individual as a separate task and leveraging shared information across tasks. However, most existing MTL methods are not explicitly designed for personalization, which makes them less effective in capturing individual-specific patterns within heterogeneous populations. This challenge is further exacerbated in high-dimensional (HD) settings, which are common in modern applications involving complex data, such as sensor signals and medical imaging. In these cases, the large parameter space and complex dependencies make it challenging to capture shared structures across individuals, which results in an increased risk of overfitting and hinders effective personalization. Tensor decomposition techniques address these challenges by identifying low-dimensional representations within HD parameter spaces \citep{kolda2009tensor}. Building on this idea, tensor regression embeds a low-rank factorization directly into the coefficient tensor of a generalized linear model \citep{zhou2013tensor, konyar2024federated, konyar2025robust}. For instance, \citep{gahrooei2021multiple} employed Tucker decomposition on the tensor of parameters of a regression model. These methods leverage the inherent low-rank structure in the data or parameter spaces to capture both common patterns and variations across subpopulations. This is particularly useful in applications such as telemonitoring, where each patient is modeled as a separate task to learn shared health trends while accurately representing individual physiological patterns. Similarly, in neuroimaging applications, tensor decomposition facilitates the extraction of personalized biomarkers from high-dimensional functional MRI data for diagnosing conditions such as attention-deficit/hyperactivity disorder (ADHD). 

\begin{figure}
        \caption{An overview of the proposed framework: (a) The individuals' model parameters are combined into a tensor that is decomposed using a Tucker decomposition. The core tensor and the factor matrices of the decomposition are estimated by solving an optimization problem; (b) the Tucker decomposition can capture the cross-correlation structures between both features and individuals.}
        \centering
    \begin{subfigure}[b]{0.65\linewidth}
        \centering
        \includegraphics[width=0.9\linewidth]{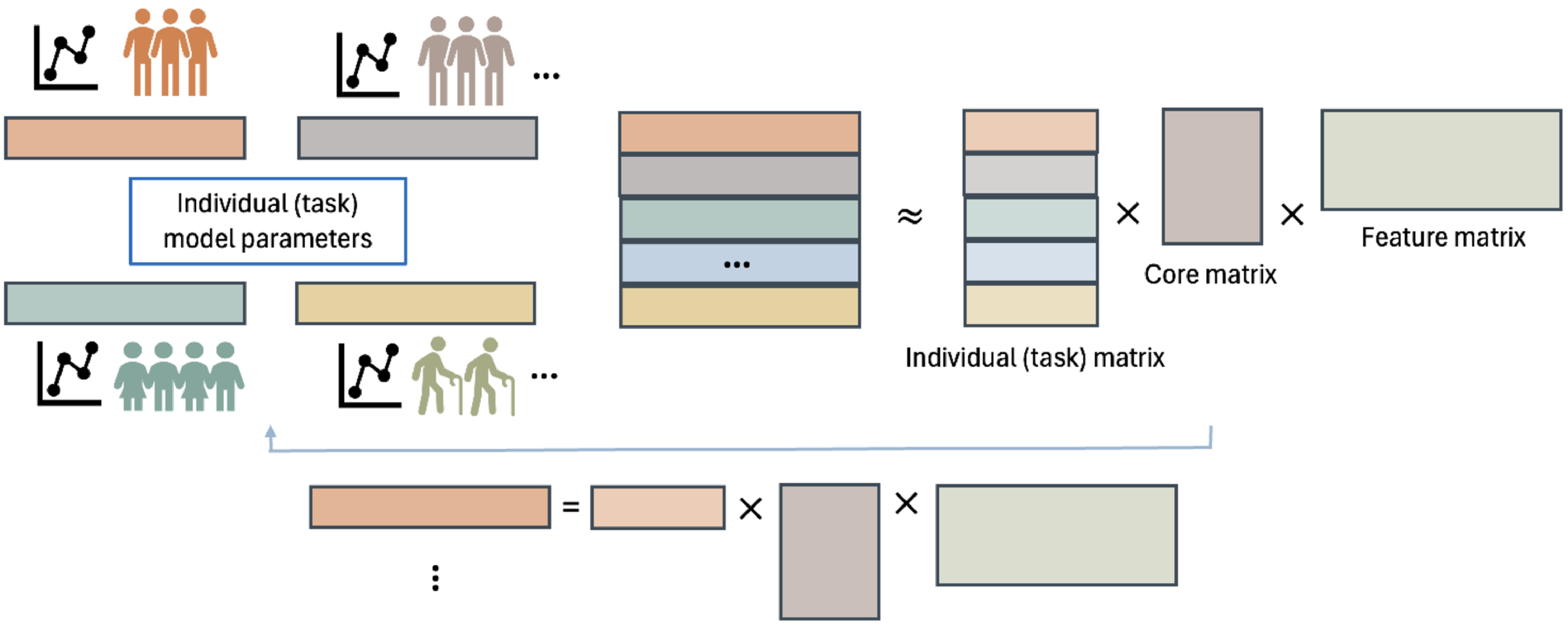}
        \caption{}
        \label{fig:overview}
    \end{subfigure}
    \hfill
    \begin{subfigure}[b]{0.33\linewidth}
        \centering
        \includegraphics[width=0.9
        \linewidth]{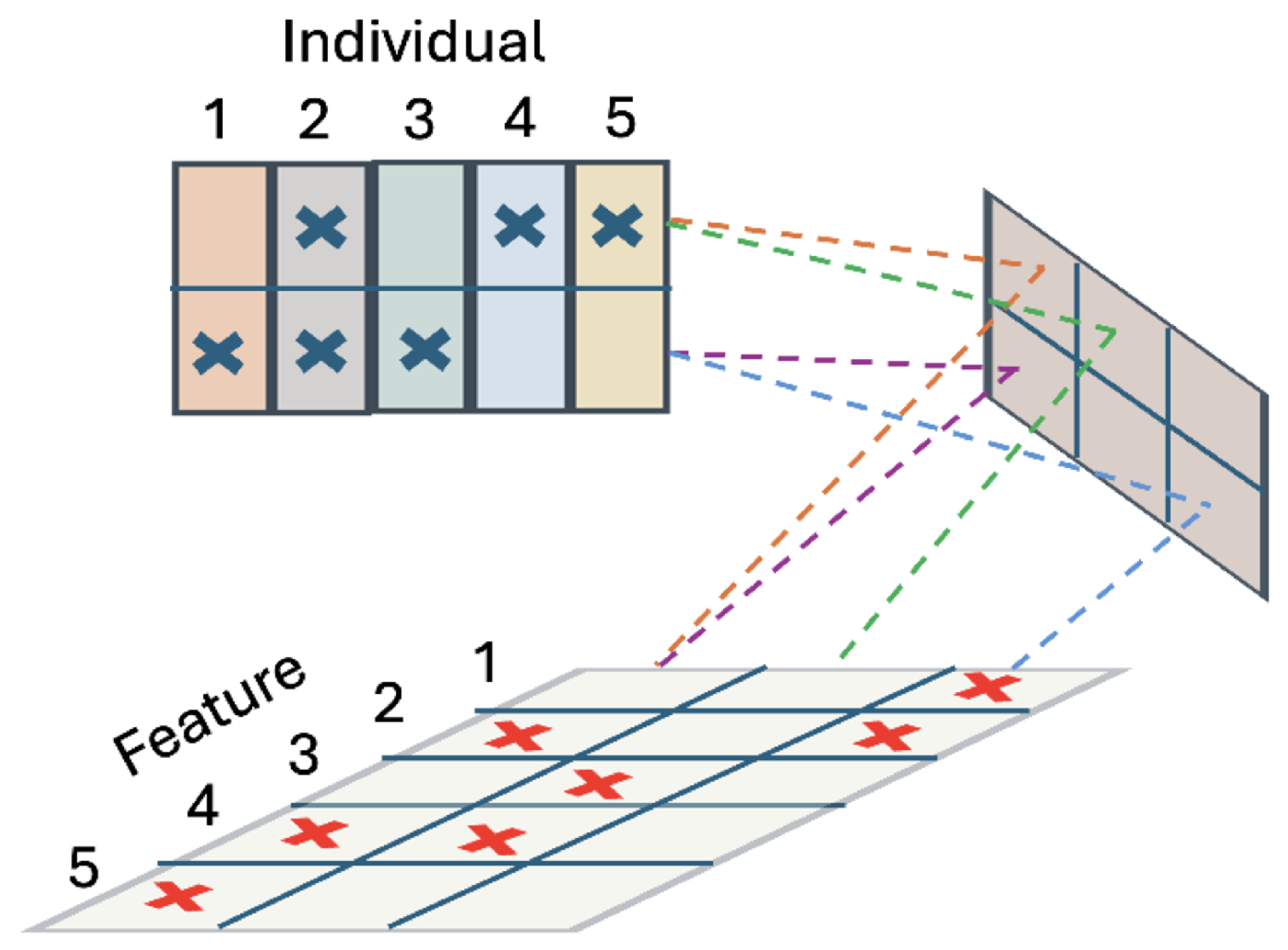}
        \caption{}
        \label{fig:corrvis}
    \end{subfigure}
    \label{fig:overviewboth}
\end{figure}

In this paper, we introduce Tensorized Multi-Task Learning (TenMTL), a framework that utilizes tensor decomposition to enhance personalized modeling across heterogeneous subpopulations. TenMTL is built upon a generalized tensor regression model designed to handle high-dimensional input data, such as functional magnetic resonance imaging (fMRI), by exploiting the multi-way structure of the data. In addition, we consider a special case where the model adopts a generalized linear model (GLM) formulation, allowing it to accommodate various outcome types. The proposed method leverages the low-rank structure of model parameters across similar individuals (or tasks) through a Tucker tensor decomposition. We formulate TenMTL as an optimization problem, where the objective function integrates several components: the minimization of the negative log-likelihood of the training data using a generalized tensor regression model, along with sparsity-inducing penalties and low-rankness constraints. Fig. \ref{fig:overviewboth} illustrates an overview of the proposed method. 

The rest of the article is organized as follows: In Section \ref{sec:prelim}, we provide preliminaries and notation. Next, we describe the proposed methodology in detail, discuss model parameter estimation algorithms, and provide a discussion in Section \ref{sec:method}. We evaluate the performance of the proposed method with three simulation scenarios in Section \ref{sec:simulation}. In Section \ref{sec:case}, we apply the proposed method and several benchmarks to two real data sets from healthcare applications to validate the performance of the proposed approach. Lastly, we conclude in Section \ref{sec:conclusion}. 

\section{Preliminaries and Tensor Notation}
\label{sec:prelim}

In this section, we introduce our notation and review some tensor algebra concepts \citep{kolda2009tensor}. We use lowercase letters for scalars, e.g., x, bold lowercase letters for vectors, e.g., $\mathbf{x}$, bold capital letters for matrices, e.g., $\mathbf{X}$, and capital calligraphic letters for tensors, e.g., $\mathcal{X}$. For instance, an order-$n$ regular tensor is denoted by $\mathcal{X} \in \mathbb{R}^{I_1 \times I_2 \ldots \times I_n}$ where the dimensions of each mode $i$ is $I_i$ ($i=1,\ldots,n$). Furthermore, $\mathcal{X}_{(j)} \in \mathbb{R}^{I_j\times I_1\ldots I_{j-1} I_{j+1} \ldots I_n}$ denotes the mode-$j$ matricization of a tensor $\mathcal{X}$ with each column corresponding to mode-$j$ fibers (tensor subarrays with all but the $j^{th}$ index fixed) of $\mathcal{X}$. Additionally,  $\text{Tr}(\mathbf{X})$ stands for the trace of a matrix $\mathbf{X}$. 

The Frobenius norm of a tensor $\mathcal{X}$ is defined as the Frobenius norm of any matricized tensor, for example, $\vert\vert\mathcal{X}\vert\vert^2_F = \vert\vert\mathbf{X}_{(1)}\vert\vert^2_F$. The mode-$j$ product of a tensor $\mathcal{X}$ by a matrix $\mathbf{A} \in \mathbb{R}^{L \times I_j}$ is a tensor in $\mathbb{R}^{I_1\times I_2\times\ldots I_{j-1} \times L \times I_{j+1} \times \ldots \times I_n}$ and is defined as $(\mathcal{X} \times_j \mathbf{A})_{i_1, i_2, \ldots i_{j-1}, l, i_{j+1}, \ldots, i_n} = \sum_{i_j = 1}^{I_j} \mathcal{X}_{i_1, \ldots, i_j, \ldots, i_n} \mathbf{A}_{l,i_j}$. The Tucker decomposition of a tensor $\mathcal{X}$ decomposes the tensor into a core tensor $\mathcal{C} \in \mathbb{R}^{P_1 \times P_2 \times \ldots \times P_n}$ and $n$ orthogonal matrices $\mathbf{U}_i \in \mathbb{R}^{I_i \times P_i} (i = 1, \ldots n)$. The decomposition can be written as $\mathcal{X} = \mathcal{C} \times_1 \mathbf{U}_1 \times_2 \mathbf{U}_2 \times_3 \ldots \times_n \mathbf{U}_n $. The dimensions of the core tensor $\mathcal{C}$ is smaller than the dimensions of $\mathcal{X}$, that is, $P_j \leq I_j (j = 1, 2, \ldots, n)$. Furthermore, $\mathbf{X} \otimes \mathbf{Y} \in \mathbb{R}^{km\times ln}$ denotes $\textit{Kronecker product}$ of two matrices, $\mathbf{X} \in \mathbb{R}^{k\times l}$ and $\mathbf{Y} \in \mathbb{R}^{m \times n}$. 

\section{Tensorized Multi-Task Learning for Model Personalization}
\label{sec:method}

In this section, we present the proposed framework, Tensorized Multi-Task Learning (TenMTL), which enables effective personalized modeling of heterogeneous subpopulations with high-dimensional data. In particular, we assume that each subpopulation (individual) has access to a limited data set containing scalar and tensor inputs and scalar outputs. TenMTL characterizes the learning task of each individual by a tensor regression model and captures common information and the similarity between individuals by imposing low-rankness on the tensor containing all individuals' model parameters. 

Assume data from $N$ individuals (subpopulations) are available. Each individual $i$ has access to a set of training data $\mathbb{D}_i=\{y_{ij}, \mathbf{z}_{ij}, \mathcal{X}_{ij} \}_{j=1}^{n_i}$, where $n_i$ is the sample size in the $i^{th}$ task (training a model for the $i^{th}$ individual), $y_{ij} \in \mathbb{R}$ is a scalar response variable, $\mathbf{z}_{ij} \in \mathbb{R}^p$ is a vector of scalar predictors, and $\mathcal{X}_{ij} \in \mathbb{R}^{I_1 \times I_2 \times \ldots \times I_m}$ is an $m$-dimensional tensor predictor. We assume the response conditioned on the predictors belongs to an exponential family distribution. That is, $y_{ij}|\mathbf{z}_{ij}, \mathcal{X}_{ij}\sim f(\theta_{ij})$, where $f(\theta_{ij}) =\exp\left(y_{ij}\theta_{ij} - b(\theta_{ij})\right)$. Then, we assume the model related to the $i^{th}$ task has a generalized linear tensor form as follows:
$g(E[y_{ij}|\mathcal{X}_{ij}, \mathbf{z}_{ij}])=\theta_{ij}= \boldsymbol{\gamma}_{i}^\top \mathbf{z}_{ij} + \left\langle \mathcal{B}_{i},\mathcal{X}_{ij}\right\rangle +e_{ij};\; j=1,2, \ldots,n_i; \; i=1,2,\ldots,N,$
where $g(.)$ is a strictly increasing link function, $b(.)$ is a distribution-specific known function, $\theta_{ij}$ is the canonical parameter of the corresponding exponential family related to $i^{th}$ individual (task), $e_{ij}$ is independent and identically distributed random error, $\boldsymbol{\gamma}_{i} \in \mathbb{R}^{p}$ is the vector of true regression coefficients for scalar predictors, and $\mathcal{B}_{i}\in \mathbb{R}^{I_{1}\times I_{2} \times \ldots\times I_{m}}$ is the true tensor of parameters. The goal is to estimate $\boldsymbol{\gamma}_{i}$ and  $\mathcal{B}_{i}$ for each $i$, while leveraging shared structures across individuals. To construct personalized models for heterogeneous subpopulations, let $\Tilde{\mathcal{B}} = \{\mathcal{B}_i\}_{i=1}^N$ denote the concatenation of $\mathcal{B}_i$, ($i=1, \ldots, N$) and $\boldsymbol{\Gamma}=[\boldsymbol{\gamma}_1, \ldots, \boldsymbol{\gamma}_N]^\top$ denote the concatenation of $\boldsymbol{\gamma}_i$, $(i=1, \ldots, N)$ along the individuals mode. Consequently, $\tilde{\mathcal{B}} \in \mathbb{R}^{N \times I_1 \times I_2 \times \ldots \times I_m}$ forms an $m+1$ dimensional tensor and $\boldsymbol{\Gamma}$ forms an $N \times p$ matrix, both of which should assume a low-rank structure due to the similarities among tasks. To impose low-rankness, \cite{li2018tucker} applied the Tucker decomposition operation (also known as higher-order singular value decomposition (HOSVD)) on the model parameters. In Tucker tensor regression models, $\mathcal{B}$ assumes the Tucker decomposition $\mathcal{B} = \sum_{r_1 = 1}^{R_1} \ldots \sum_{r_m = 1}^{R_m} g_{r_1, \ldots, r_m} \boldsymbol{u}_1^{(r_1)} \circ \ldots \circ \boldsymbol{u}_m^{(r_m)}$, where $\boldsymbol{u}_d^{(r_d)} \in \mathbb{R}^{I_d}$ ($d=1, \ldots, m$; $r_d = 1, \ldots, R_d$) are all column vectors, and $g_{r_1, \ldots, r_m}$ are constants. The decomposition can also be denoted as $\mathcal{B} = \llbracket \mathcal{G} ; \mathbf{U}_1, \ldots, \mathbf{U}_m \rrbracket$, where $\mathcal{G} \in \mathbb{R}^{R_1 \times \ldots \times R_m}$ is the $m$-dimensional core tensor with $g_{r_1, \ldots, r_m}$ being $(r_1, \ldots, r_m)^{th}$ entry, and $\mathbf{U}_d = [\boldsymbol{u}_d^{(1)}, \ldots,  \boldsymbol{u}_d^{(R_d)}] \in \mathbb{R}^{I_d \times R_d}$ are the factor matrices.

In the proposed TenMTL framework, we impose a Tucker decomposition form on the tensor of concatenated model parameters $\tilde{\mathcal{B}}$ and $\boldsymbol{\Gamma}$. That is, $\tilde{\mathcal{B}} = [\![\mathcal{G}; \mathbf{U}_{0}, \mathbf{U}_1, \ldots, \mathbf{U}_m]\!]$, where $\mathbf{U}_{0} \in \mathbb{R}^{N \times R_{0}}$ is the factor matrix capturing correlations between tasks (individual models), $\mathbf{U}_d \in \mathbb{R}^{I_d \times R_d}$, $(d=1,2,\ldots,m)$ are the factor matrices along each mode $d$, and $\mathcal{G} \in \mathbb{R}^{R_{0} \times R_1 \times \ldots \times R_m}$ is the core tensor. Similarly, $\boldsymbol{\Gamma} = [\![\mathcal{H}; \mathbf{V}_{0}, \mathbf{V}_1 ]\!]$, where $\mathbf{V}_{0} \in \mathbb{R}^{N \times T_0}$ is the factor matrix capturing correlations between tasks (individual models) through vector parameters, $\mathbf{V}_1 \in \mathbb{R}^{p \times T_1}$ is the factor matrix along the feature mode, and $\mathcal{H} \in \mathbb{R}^{T_0 \times T_1}$ is the core tensor. We assume that the factor matrices associated with feature modes, i.e., $\mathbf{U}_1, \ldots, \mathbf{U}_m, \mathbf{V}_1$, are shared across individuals (subgroups). Moreover, we assume that the factor matrices related to the individuals' mode, $\mathbf{U}_0$ and $\mathbf{V}_0$, have a set of common columns and a set of distinct ones. More specifically, we assume $\mathbf{U}_0 = [\mathbf{W}_0, \mathbf{F}_0]$ and $\mathbf{V}_0 = [\mathbf{W}_0, \mathbf{D}_0]$, where $\mathbf{W}_0 \in \mathbb{R}^{N \times Q_0}$ characterizes the similarities that are captured by both tensor and scalar input data, $\mathbf{F}_0 \in \mathbb{R}^{N \times (R_0 - Q_0)}$ are the components capturing correlations between tasks through tensor data, and $\mathbf{D}_0 \in \mathbb{R}^{N \times (T_0 - Q_0)}$ are the components capturing correlations between tasks through scalar inputs. 
Furthermore, we allow subgroups to personalize their models through $\mathbf{U}_0$ and $\mathbf{V}_0$. In some cases, subpopulations have correlations through a subset of features. For example, in health sciences, only some subpopulations may have encountered a specific disease due to their environmental conditions \citep{weiss2004social}. To address subgroup-specific correlation through a subset of features, we add a lasso penalty to $\mathcal{G}$, $\mathcal{H}$, $\{\mathbf{U}_d\}$, and $\mathbf{V}_1$. Adding a lasso penalty also helps with having numerical stability. Then, we formulate the following problem that minimizes the negative log-likelihood to estimate the personalized model parameters of each individual:
\begin{equation}
\label{eq:tensoreq2}
    \begin{aligned}
        &\min_{\gamma_{i}, \mathcal{B}_i } \; \sum_{i=1}^N \sum_{j=1}^{n_i} \big( -y_{ij} (\boldsymbol{\gamma}_i^\top \mathbf{z}_{ij} + \langle \mathcal{B}_i, \mathcal{X}_{ij} \rangle) + b(\boldsymbol{\gamma}_i^\top \mathbf{z}_{ij} +  \langle \mathcal{B}_i, \mathcal{X}_{ij} \rangle) \big) +\lambda_g \vert\vert \mathcal{G}\vert\vert_1 + \lambda_h \vert\vert \mathcal{H}\vert\vert_1 \\& \qquad+ \sum_{d=1}^m \lambda_u \vert\vert \mathbf{U}_d \vert\vert_1 + \lambda_v \vert\vert \mathbf{V}_1 \vert\vert_1\\
        &\;\; \emph{s.t.} \;\quad \Tilde{\mathcal{B}}= \{\mathcal{B}_i \}_{i=1}^N, \\
        &\qquad\quad\;\boldsymbol{\Gamma}=[\boldsymbol{\gamma}_1, \ldots, \boldsymbol{\gamma}_N]^\top, \\
        &\qquad\quad\; \Tilde{\mathcal{B}} = [\![\mathcal{G}, \mathbf{U}_0, \mathbf{U}_1, \ldots, \mathbf{U}_m]\!], \\
        &\qquad\quad\; \boldsymbol{\Gamma} = [\![\mathcal{H}, \mathbf{V}_0, \mathbf{V}_1]\!], \\
        &\qquad\quad\; \mathbf{U}_0 = [\mathbf{W}_0, \mathbf{F}_0], \\
        &\qquad\quad\; \mathbf{V}_0 = [\mathbf{W}_0, \mathbf{D}_0],
    \end{aligned}
\end{equation}
where $\lambda_g \geq 0$, $\lambda_h \geq 0$, $\lambda_u \geq 0$, and $\lambda_v \geq 0$ are scalar hyperparameters. To solve this problem, we incorporate the block coordinate descent algorithm. That is, we solve the problem with respect to each variable (i.e., $\mathbf{W}_0, \mathbf{F}_0, \mathcal{G}, \{\mathbf{U}_{d}\}, \mathbf{D}_0, \mathcal{H}, \mathbf{V}_1$) sequentially by fixing the remaining variables. Next, we discuss the update strategies for the model parameters: \\
\noindent\textbf{Update of \texorpdfstring{$\mathbf{W}_0$}{Lg}: }
Let $\Tilde{\mathcal{B}}_{(0)}$, $\mathcal{G}_{(0)}$, $\boldsymbol{\Gamma}_{(0)}$, and $\mathcal{H}_{(0)}$ be the matricizations of $\Tilde{\mathcal{B}}$, $\mathcal{G}$, $\boldsymbol{\Gamma}$, and $\mathcal{H}$ along the \textit{task} mode. Then, we rewrite the decompositions as $\Tilde{\mathcal{B}}_{(0)} = \mathbf{U}_0 \mathcal{G}_{(0)} (\mathbf{U}_m \otimes \ldots \otimes \mathbf{U}_1)^\top $, where $\Tilde{\mathcal{B}}_{(0)} \in \mathbb{R}^{N \times I_1 I_2 \ldots I_m}$ and $\mathcal{G}_{(0)} \in \mathbb{R}^{R_0 \times R_1 R_2 \ldots R_m}$, and $\boldsymbol{\Gamma}_{(0)} = \mathbf{V}_0 \mathcal{H}_{(0)} \mathbf{V}_1^\top$, where $\boldsymbol{\Gamma}_{(0)} \in \mathbb{R}^{N \times p}$ and $\mathcal{H}_{(0)} \in \mathbb{R}^{T_0 \times T_1}$. Let $\mathbf{K}_0 := \mathbf{U}_m \otimes \ldots \otimes \mathbf{U}_1 \in \mathbb{R}^{I_1\ldots I_m \times R_1 \ldots R_m}$. Then, TenMTL solves the following problem to update each row of $\mathbf{W}_0$, denoted by $\mathbf{w}_{0_i} \in \mathbb{R}^{1 \times Q_0}$, as follows:
\begin{equation}
\label{eq:tensoruind}
    \begin{aligned}
        &\min_{\mathbf{w}_{0_i} } \; \sum_{j=1}^{n_i} ( -y_{ij} (\mathbf{v}_{0_i}\mathcal{H}_{(0)} \mathbf{V}_1^\top  \mathbf{z}_{ij} +\langle \mathbf{u}_{0_i} \mathcal{G}_{(0)} \mathbf{K}_0^\top,  \text{vec}(\mathcal{X}_{ij}) \rangle) + b(\mathbf{v}_{0_i}\mathcal{H}_{(0)} \mathbf{V}_1^\top \mathbf{z}_{ij} + \langle \mathbf{u}_{0_i} \mathcal{G}_{(0)} \mathbf{K}_0^\top,  \text{vec}(\mathcal{X}_{ij}) \rangle) ), \\
        &\;\; \emph{s.t.} \;\quad \mathbf{u}_{0_i} = [\mathbf{w}_{0_i}, \mathbf{f}_{0_i}], \\
        &\qquad\quad\; \mathbf{v}_{0_i} = [\mathbf{w}_{0_i}, \mathbf{d}_{0_i}].
    \end{aligned}
\end{equation}
Let $\mathbf{o}_{ij}^\top:=\text{vec}(\mathcal{X}_{ij}) \mathbf{K}_0 \mathcal{G}_{(0)}^\top \in \mathbb{R}^{1 \times R_0}$, $\mathbf{k}_{ij}:=\mathcal{H}_{(0)} \mathbf{V}_1^\top  \mathbf{z}_{ij} \in \mathbb{R}^{T_0 \times 1}$, $\mathbf{o}_{ij} = [\mathbf{o}^{(1)}_{ij}, \mathbf{o}^{(2)}_{ij}]$, $\mathbf{k}_{ij} = [\mathbf{k}^{(1)}_{ij}, \mathbf{k}^{(2)}_{ij}]$, $\psi_{ij} = \mathbf{d}_{0_i}\mathbf{k}^{(2)}_{ij}$, and $\xi_{ij}=\mathbf{f}_{0_i}\mathbf{o}^{(2)}_{ij}$. 
By using the properties of the inner product, \eqref{eq:tensoruind} can be converted into the objective of a generalized linear model (GLM) to be solved to update $\mathbf{w}_{0_i}$ for each $i$ as follows:
\begin{equation}
\label{eq:tensoruind22f}
    \begin{aligned}
        &\min_{\mathbf{w}_{0_i} } \; \sum_{j=1}^{n_i} ( -y_{ij} (\mathbf{w}_{0_i}\mathbf{k}^{(1)}_{ij} + \psi_{ij} + \mathbf{w}_{0_i}\mathbf{o}^{(1)}_{ij} + \xi_{ij} ) + b(\mathbf{w}_{0_i}\mathbf{k}^{(1)}_{ij} + \psi_{ij} + \mathbf{w}_{0_i}\mathbf{o}^{(1)}_{ij} + \xi_{ij} ) ).
    \end{aligned}
\end{equation}

\noindent\textbf{Update of \texorpdfstring{$\mathbf{F}_0$}{Lg}: }
TenMTL solves the following problem to update each row of $\mathbf{F}_0$, denoted by $\mathbf{f}_{0_i} \in \mathbb{R}^{1 \times (R_0-Q_0)}$, as follows:
\begin{equation}
\label{eq:tensoruindnewf}
    \begin{aligned}
        &\min_{\mathbf{f}_{0_i} } \; \sum_{j=1}^{n_i} ( -y_{ij} (\delta_{ij} + \psi_{ij} + \omega_{ij} + \mathbf{f}_{0_i}\mathbf{o}^{(2)}_{ij} ) + b(\delta_{ij} + \psi_{ij} + \omega_{ij} + \mathbf{f}_{0_i}\mathbf{o}^{(2)}_{ij} ) ),
    \end{aligned}
\end{equation}
where $\delta_{ij} = \mathbf{w}_{0_i}\mathbf{k}^{(1)}_{ij}$ and $\omega_{ij} = \mathbf{w}_{0_i}\mathbf{o}^{(1)}_{ij}$. Next, $\mathbf{U}_0$ is updated as $[\mathbf{W}_0, \mathbf{F}_0]$.

\noindent\textbf{Update of \texorpdfstring{$\mathbf{U}_d$}{Lg}: } 
Let $\mathcal{D}_i := \mathcal{G} \times_1 \mathbf{u}_{0_i}\times_2 \mathbf{U}_1 \ldots \times_{m+1} \mathbf{U}_m \in \mathbb{R}^{1 \times I_1 \times \ldots \times I_m}$. Then, $\mathcal{D}_{i{(d)}}=\mathbf{U}_{d} \mathcal{G}_{(d)} (\mathbf{U}_m \otimes \ldots \mathbf{U}_{d+1} \otimes \mathbf{U}_{d-1} \otimes \ldots \mathbf{U}_1 \otimes \mathbf{u}_{0_i})^\top$ denotes the mode-$d$ matricization of $\mathcal{D}_i$. To update $\mathbf{U}_{d}$, TenMTL solves the following problem:
\begin{equation}
\label{eq:tensorud2}
    \begin{aligned}
        &\min_{\mathbf{U}_{d} } \; \sum_{i=1}^N \sum_{j=1}^{n_i} ( -y_{ij} (\boldsymbol{\gamma}_i^\top \mathbf{z}_{ij} + \langle \mathbf{U}_{d} \mathcal{G}_{(d)} \mathbf{K}_{d_i}^\top,  \mathcal{X}_{ij_{(d)}} \rangle) + b(\boldsymbol{\gamma}_i^\top \mathbf{z}_{ij} + \langle \mathbf{U}_{d} \mathcal{G}_{(d)} \mathbf{K}_{d_i}^\top, \mathcal{X}_{ij_{(d)}} \rangle) ) + \lambda_u \vert\vert \mathbf{U}_d \vert\vert_1,
    \end{aligned}
\end{equation}
where $\mathbf{K}_{d_i}:=\mathbf{U}_m \otimes \ldots \mathbf{U}_{d+1} \otimes \mathbf{U}_{d-1} \otimes \ldots \mathbf{U}_1 \otimes \mathbf{u}_{0_i}$. 
By letting $\mathbf{u}_d:=\text{vec}(\mathbf{U}_{d})$ and $\mathbf{s}_{ij}^\top:=\text{vec}(\mathcal{X}_{ij_{(d)}}\mathbf{K}_{d_i}\mathcal{G}_{(d)}^\top)$, the following lasso regularized GLM is solved to update $\mathbf{u}_d$:
\begin{equation}
\label{eq:tensorudupd}
    \begin{aligned}
        &\min_{\mathbf{u}_d} \; \sum_{i=1}^N \sum_{j=1}^{n_i} ( -y_{ij} (\boldsymbol{\gamma}_i^\top \mathbf{z}_{ij} + \mathbf{u}_d \mathbf{s}_{ij} ) + b(\boldsymbol{\gamma}_i^\top \mathbf{z}_{ij} + \mathbf{u}_d \mathbf{s}_{ij}) ) +\lambda_u \vert\vert \mathbf{u}_d \vert\vert_1.
    \end{aligned}
\end{equation}
\noindent\textbf{Update of \texorpdfstring{$\mathcal{G}$}{Lg}: } 
TenMTL solves the following problem to update $\mathcal{G}$:
\begin{equation}
\label{eq:tensorG}
    \begin{aligned}
        \min_{\mathcal{G}_{(0)} } \; &\sum_{i=1}^N \sum_{j=1}^{n_i} ( -y_{ij} (\boldsymbol{\gamma}_i^\top \mathbf{z}_{ij} + \langle \mathcal{G}_{(0)} ,  \mathbf{u}_{0_i}^\top\text{vec}(\mathcal{X}_{ij})\mathbf{K}_0 \rangle)  + b(\boldsymbol{\gamma}_i^\top \mathbf{z}_{ij} + \langle  \mathcal{G}_{(0)},  \mathbf{u}_{0_i}^\top\text{vec}(\mathcal{X}_{ij})\mathbf{K}_0 \rangle) ) + \lambda_g \vert\vert \mathcal{G}_{(0)} \vert\vert_1.
    \end{aligned}
\end{equation}
by setting $\mathbf{g}:= \text{vec}(\mathcal{G}_{(0)}) \in \mathbb{R}^{1 \times R_0R_1 \ldots R_m}$ and $\mathbf{w}_{ij}^\top:=\text{vec}(\mathbf{u}_{0_i}^\top\text{vec}(\mathcal{X}_{ij})\mathbf{K}_0) \in \mathbb{R}^{1 \times R_0R_1 \ldots R_m}$, the following lasso regularized GLM is solved to update $\mathbf{g}$:
\begin{equation}
\label{eq:tensorG2}
    \begin{aligned}
        &\min_{\mathbf{g} } \; \sum_{i=1}^N \sum_{j=1}^{n_i} ( -y_{ij} (\boldsymbol{\gamma}^\top \mathbf{z}_{ij} + \mathbf{g}  \mathbf{w}_{ij} + b(\boldsymbol{\gamma}^\top \mathbf{z}_{ij} + \mathbf{g}  \mathbf{w}_{ij}) )  + \lambda_g \vert\vert \mathbf{g} \vert\vert_1.
    \end{aligned}
\end{equation}

\noindent\textbf{Update of \texorpdfstring{$\mathbf{D}_0$}{Lg}: } 
To update $\mathbf{D}_0$, we solve the following generalized linear model for $\mathbf{d}_{0_i}$ for all individuals:
\begin{equation}
\label{eq:tensoruind22f22}
    \begin{aligned}
        &\min_{\mathbf{d}_{0_i} } \; \sum_{j=1}^{n_i} ( -y_{ij} (\delta_{ij} + \mathbf{d}_{0_i}\mathbf{k}^{(2)}_{ij} + \omega_{ij} + \epsilon_{ij} ) + b(\delta_{ij} + \mathbf{d}_{0_i}\mathbf{k}^{(2)}_{ij} + \omega_{ij} + \epsilon_{ij} ) ).
    \end{aligned}
\end{equation}
Next, we update $\mathbf{V}_0$ as $\mathbf{V}_0 = [\mathbf{W}_0, \mathbf{D}_0]$.

\noindent\textbf{Update of \texorpdfstring{$\mathbf{V}_1$}{Lg}: } 
We solve the following problem:
\begin{equation}
\label{eq:tensoruindV}
    \begin{aligned}
        &\min_{\mathbf{V}_1 } \; \sum_{i=1}^N \sum_{j=1}^{n_i} ( -y_{ij} (\boldsymbol{\kappa}_i \mathbf{V}_1^\top  \mathbf{z}_{ij} +\hat{a}_{ij} ) + b(\boldsymbol{\kappa}_i \mathbf{V}_1^\top \mathbf{z}_{ij} +\hat{a}_{ij}) ) + \lambda_v \vert\vert \mathbf{V}_1 \vert\vert_1,
    \end{aligned}
\end{equation}
where $\hat{a}_{ij}=\langle \mathbf{u}_{0_i} \mathcal{G}_{(0)} \mathbf{K}_0^\top,  \text{vec}(\mathcal{X}_{ij}) \rangle$ and $\boldsymbol{\kappa}_i:=\mathbf{v}_{0_i}\mathcal{H}_{(0)} \in \mathbb{R}^{1 \times T_1}$. By setting $\mathbf{v}_1:=\text{vec}(\mathbf{V}_{1}^\top) \in \mathbb{R}^{T_1p}$ and $\mathbf{g}_{ij}:=\mathbf{z}_{ij}^\top \otimes \boldsymbol{\kappa}_i \in \mathbb{R}^{1 \times T_1p}$, we solve the following lasso regularized generalized linear model (GLM):
\begin{equation}
\label{eq:updateV2}
\begin{aligned}
        &\min_{\mathbf{v}_1  } \; \sum_{i=1}^N \sum_{j=1}^{n_i} \left( -y_{ij}(\mathbf{g}_{ij}\mathbf{v}_1 + \hat{a}_{ij})+ b(\mathbf{g}_{ij}\mathbf{v}_1 + \hat{a}_{ij}) \right) +\lambda_v\vert\vert \mathbf{v}_1 \vert\vert_1.
        \end{aligned}
\end{equation}

\noindent\textbf{Update of \texorpdfstring{$\mathcal{H}$}{Lg}: } 
Let $\mathbf{b}_{ij}:=\mathbf{V}_1^\top \mathbf{z}_{ij} \in \mathbb{R}^{T_1 \times 1}$. We then transform and solve the following lasso-regularized GLM:
\begin{equation}
\label{eq:updateH}
\begin{aligned}
        &\min_{\mathbf{h}} \; \sum_{i=1}^N \sum_{j=1}^{n_i} ( -y_{ij}(\mathbf{c}_{ij} \mathbf{h} +\hat{a}_{ij}) + b(\mathbf{c}_{ij} \mathbf{h}+\hat{a}_{ij} ) ) +\lambda_h \vert\vert \mathbf{h} \vert\vert_1,
\end{aligned}
\end{equation}
where $\mathbf{h}=\text{vec}(\mathcal{H}_{(0)}) \in \mathbb{R}^{T_0T_1}$, and $\mathbf{c}_{ij}=\mathbf{b}_{ij}^\top \otimes \mathbf{v}_{0_i} \in \mathbb{R}^{1 \times T_0T_1}$.

The update procedures for $\mathbf{W}_0$, $\mathbf{F}_0$, $\mathbf{U}_d$, $\mathcal{G}$, $\mathbf{D}_0$, $\mathbf{V}_1$, and $\mathcal{H}$  are repeated until convergence, i.e., $\frac{l^t - l^{t-1}}{l^{t-1}} < \epsilon$ where $l^t$ is the objective function value at iteration $t$. After estimating $\{\mathbf{u}_{0_i}\}_{i=1}^N$, $\mathcal{G}$, and $\{\mathbf{U}_{d}\}_{d=1}^m$, $\{\mathbf{v}_{0_i}\}_{i=1}^N$, $\mathcal{H}$, and $\mathbf{V}_{1}$, the parameters for personalized models can be obtained by $\mathcal{B}_i = \mathcal{G} \times_1 \mathbf{u}_{0_i} \times_2 \mathbf{U}_{1}  \ldots \times_{m+1} \mathbf{U}_{m}$ and $\boldsymbol{\gamma}_i = \mathcal{H} \times_1 \mathbf{v}_{0_i} \times_2 \mathbf{V}_{1}$. Note that the resulting models contain individual-specific information through $\mathbf{U}_0$ and $\mathbf{V}_0$, along with shared knowledge through $\mathcal{G}$, $\mathcal{H}$, $\{\mathbf{U}_{d}\}$, and $\mathbf{V}_1$. Thus, personalization is achieved by fine-tuning the shared knowledge using the individual-specific vectors $\mathbf{u}_{0_i}$ and $\mathbf{v}_{0_i}$. A clear benefit of framing the parameter estimation problem as an optimization framework is its flexibility to include prior knowledge, any structural constraints we might wish to apply to the models. Furthermore, TenMTL offers flexibility in handling vector data, tensor data, and their combinations, which makes it well-suited for multi-modal data modeling. In the special case where only vector data is available, the framework reduces to a set of lasso-regularized generalized linear models (GLMs). Details of the parameter estimation for this special case are provided in Appendix.

\begin{algorithm}
\caption{Tensor-based Multi-Task Learning for Personalization}
\begin{algorithmic}[1]   
\Procedure{TenMTL}{$\mathcal{X}_{ij}, \mathbf{z}_{ij}, y_{ij}, \lambda_g, \lambda_h, \lambda_u, \lambda_v$}
    \State Initialize $\mathbf{W}_0, \mathbf{F}_0, \mathbf{U}_d, \mathcal{G}, \mathbf{D}_0, \mathbf{V}_1, \mathcal{H}$ by applying Tucker decomposition to the concatenation of locally estimated parameters from the generalized tensor regression models.
    \While{convergence criterion not met}
        \State Update $\mathbf{W}_0$ with Eq.~\eqref{eq:tensoruind22f}.
        \State Update $\mathbf{F}_0$ with Eq.~\eqref{eq:tensoruindnewf}.
        \State Update $\mathbf{U}_d$, $\forall d$ with Eq.~\eqref{eq:tensorud2}.
        \State Update $\mathcal{G}$ with Eq.~\eqref{eq:tensorG2}.
        \State Update $\mathbf{D}_0$ with Eq.~\eqref{eq:tensoruind22f22}.
        \State Update $\mathbf{V}_1$ with Eq.~\eqref{eq:updateV2}.
        \State Update $\mathcal{H}$ with Eq.~\eqref{eq:updateH}.
    \EndWhile
    \State \textbf{return} $\mathbf{W}_0, \mathbf{F}_0, \mathbf{U}_d, \mathcal{G}, \mathbf{D}_0, \mathbf{V}_1, \mathcal{H}$
\EndProcedure
\end{algorithmic}
\end{algorithm}

\subsection{Discussion}

In this section, we discuss the effect of rank on the model and how the proposed method can be related to the mixed-effect models. Furthermore, we provide a discussion on how knowledge can be transferred when there are new tasks. Finally, we discuss how the proposed method can be applied to multi-modal datasets.

\subsubsection{Effect of Rank on the Parameter Estimation}
\label{sec:modelarch}

Mixed-effect models are well-established in the literature for their effectiveness in modeling subpopulations with inherent heterogeneity and individual-specific characteristics \citep{laird1982random}. These models incorporate both fixed effects, which capture population-level trends, and random effects, which model individual-level variability as stochastic deviations from the mean. In contrast, the proposed method uses a deterministic low-rank decomposition to represent variation across individuals. While it may initially appear to account only for fixed effects, the low-rank structure inherently captures shared patterns and structured individual differences through factorization. Importantly, this decomposition can be reinterpreted in a stochastic framework. By assigning a probabilistic distribution to the latent individual-specific components, the proposed method can be viewed as a reduced-rank random-effects model. As detailed in the remark below, the flexibility of this framework allows it to span a continuum from fixed-effect models to full mixed-effect models by adjusting the rank of the latent space associated with individual-level variability. For illustration, we focus on the vector-based setting, although a similar result readily extends to the tensor case.

\begin{remark} \label{remark} \textbf{(Random-Effect Interpretation of the Proposed Method)} Consider the tensor factorization model for individual-specific coefficients:
\[
\boldsymbol{\beta}_i = \mathbf{u}_{0_i} \mathcal{G}_{(0)} \mathbf{U}_1^\top  = \mathbf{u}_{0_i} \mathbf{L} , 
\]
where $\mathbf{L} := \mathcal{G}_{(0)}\mathbf{U}_1^\top \in \mathbb{R}^{ R_0 \times p}$, and assume that $\mathbf{u}_{0_i} \sim \mathcal{N}(0, \mathbf{I}_{R_0})$. Here, $\mathbf{u}_{0_i} \in \mathbb{R}^{1 \times R_0}$ is a latent random vector encoding individual-level variability, $\mathcal{G}_{(0)} \in \mathbb{R}^{R_0 \times R_1}$ is the shared core matrix, and $\mathbf{U}_1 \in \mathbb{R}^{p \times R_1}$ is the shared feature basis matrix. By treating $\mathbf{u}_{0_i}$ as a random variable, depending on the value of \(R_0\), the model behaves as follows:

\begin{enumerate}
    \item[(1)] Degenerate model (\(R_0 = 0\)):  
    $
    \boldsymbol{\beta}_i = \mathbf{0}$ and $\text{Cov}(\boldsymbol{\beta}_i) = \mathbf{0}_{p \times p}.
    $
    No individual variability is captured. All individuals share the same zero coefficients.

    \item[(2)] Rank-1 random effect model (\(R_0 = 1\)):
    $\mathbf{U}_0 \in \mathbb{R}^{N \times 1}$ and $\boldsymbol{\beta_i} = \mathbf{u}_{0_i} (\mathcal{G}_{(0)} \mathbf{U}_1^\top)$, where $ \mathbf{u}_{0_i}$ is a scalar with $\mathbf{u}_{0_i} \sim \mathcal{N}(0,1) $ and $\mathbf{v} = \mathcal{G}_{(0)} \mathbf{U}_1^\top$ is a $p$-dimensional vector. In this case,
    covariance is rank-1: \(\text{Cov}(\boldsymbol{\beta}_i) = \mathbf{v} \mathbf{v}^\top\).  
    All inter-individual variation lies along a single direction in the feature space. This can be interpreted as all $\boldsymbol{\beta_i}$ lie on the same vector $\mathbf{v}$, but scaled by $\mathbf{u}_{0_i}$. 
    
    \item[(3)] Reduced-rank mixed-effect model (\(1 < R_0 < N\)):  We have $\boldsymbol{\beta}_i = \mathbf{u}_{0_i} \mathbf{L}$ and $\mathbf{u}_{0_i} \sim \mathcal{N}(0, \mathbf{I}_{R_0})$, with covariance
    $\text{Cov}(\boldsymbol{\beta}_i) = \mathbf{L} \mathbf{L}^\top = \mathcal{G}_{(0)}\mathbf{U}_1^\top  \mathbf{U}_1\mathcal{G}_{(0)}^\top$.
    The model captures structured, low-dimensional variation across individuals. The rank of the covariance is at most \(R_0\).

    \item[(4)] Full-rank mixed-effect model (\(R_0 = N\)): $\boldsymbol{\beta}_i =\mathbf{u}_{0_i} \mathbf{L}$ where $\mathbf{u}_{0_i} \sim \mathcal{N}(0, \mathbf{I}_N)$ and $\mathbf{L} \in \mathbb{R}^{N  \times p}$ with covariance:
    $\text{Cov}(\beta_i) = \mathbf{L} \mathbf{L}^\top$ with rank $\leq \min(N, R_1, p)$.
    The model allows full flexibility across individuals with no low-dimensional embedding, though feature-wise variation remains constrained if \(R_1 < p\).
\end{enumerate}
\end{remark}

Thus, the proposed method exhibits a high degree of flexibility, as it can incorporate both fixed effects and individual-level variations through careful tuning of the decomposition rank.

\subsubsection{Transferring Knowledge to New Tasks}

A key advantage of the proposed TenMTL framework lies in its ability to facilitate knowledge transfer to new tasks, aligning with the principles of transfer learning. Transfer learning seeks to improve performance on a target task by leveraging knowledge from related source tasks, particularly when the target task has limited data availability \citep{weiss2016survey}. TenMTL achieves this by modeling tasks through a shared low-rank tensor structure, which captures population-level patterns via shared latent components, i.e., the core tensors $\mathcal{G}$ and $\mathcal{H}$, and the task-invariant factor matrices $\{\mathbf{U}_d\}_{d=1}^m$ and $\mathbf{V}_1$. Meanwhile, task-specific variations are captured through individualized components $\mathbf{U}_0$ and $\mathbf{V}_0$. When a new task is introduced, only the parameters $\mathbf{u}_{0_i}$ and $\mathbf{v}_{0_i}$ need to be learned, which enables efficient adaptation. This structure allows the framework to generalize to new individuals or subgroups with reduced computational overhead and minimal labeled data.

When a new task is introduced, the shared components learned from related tasks act as prior knowledge and enable the model to generalize effectively even with limited data. This idea is closely related to in-context learning, where models use previously seen task examples to perform new tasks without changing their parameters \citep{brown2020language}. Although TenMTL updates its parameters for each new task, it still benefits from reusing shared low-dimensional components learned from other tasks, similar to how in-context learning reuses patterns. This structure allows the model to adapt efficiently by updating only a small number of task-specific parameters.

\subsubsection{Scalable Modeling for Multi-Channel and Multi-Modal Data}
\label{sec:multimodal}

Advancements in data collection technologies have led to a rapid increase in multi-modal data across domains such as healthcare and agriculture \citep{korthals2018multi, li2022multi, zhang2022heterogeneous, amini2025federated}. In healthcare, data sources include electronic health records (EHRs), medical imaging, genomics, and wearable devices; in agriculture, they include satellite imagery, IoT-based environmental sensing, and crop yield measurements. These heterogeneous modalities often offer complementary insights that support personalized medicine and sustainable farming. Effectively leveraging such data requires analytical methods capable of capturing both shared and modality-specific patterns.

\begin{figure}[H]
        \caption{Illustration for an example of multi-channel data}
    \centering
    \includegraphics[width=0.6\linewidth]{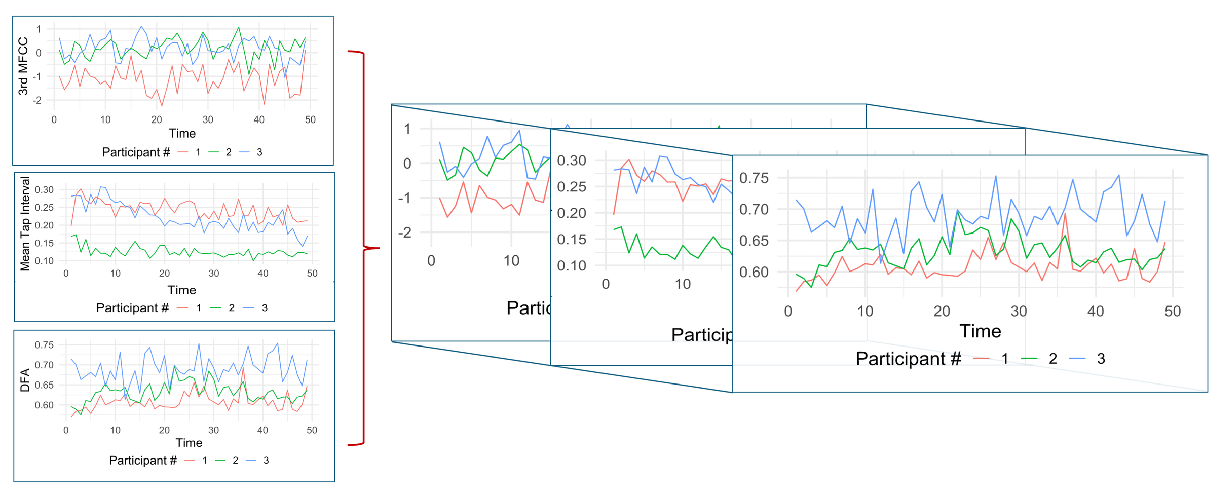}
    \label{fig:multichannel}
\end{figure}

A central strength of the proposed method lies in its adaptability to multi-modal data settings. 
For example, the proposed method naturally accommodates cases where each individual has access to multimodal datasets where all modalities have the same number of features. Examples of such situations arise when multiple sensors are collecting data over a fixed period with a given frequency, generating multi-channel datasets as it is illustrated in Figure \eqref{fig:multichannel}. In such cases, the data can be represented as a tensor, capturing both individual and interaction effects across modalities \citep{kolda2009tensor, papalexakis2016tensors}. Furthermore, the proposed method effectively models both shared and distinct structures among the scalar and tensor predictors.

Moreover, the method can be extended to general multimodal data settings where the number of features varies across modalities. The general TenMTL architecture can integrate other low-rank decompositions, such as coupled tensor decomposition (CTD), which is particularly suited for jointly analyzing datasets with shared modes \citep{acar2011all}. In this scenario, the tensor of parameters associated with each modality shares a common (or semi-common) factor matrix across the individual mode. This factor matrix will capture the individual similarities that commonly exist across all modes.
\subsection{Computational Complexity}

In this section, we study the computational complexity of the proposed method. Let us start with the computational complexity of updating $\mathbf{U}_{0}$. The computation of $\mathbf{o}_{ij}$ by using mode-wise tensor-times-matrix product instead of explicitly computing $\mathbf{K}_{0}$ reduces to $O(\sum_{d=1}^m R_1 \ldots r_{d-1} I_d R_{d+1} \ldots R_m + \Pi_{d=1}^m R_d R_{0})$. The complexity of updating $u_{0_i}$ using a generalized linear model solver in Python (e.g., \texttt{sklearn.linear\_model.LogisticRegression}) is $O(T R_{0} n_i)$, where $T$ is the number of iterations until convergence for the L-BFGS optimizer. The complexity of updating $\mathbf{U}_{0}$ becomes $O( N(\sum_{d=1}^m R_1 \ldots r_{d-1} I_d R_{d+1} \ldots R_m + \Pi_{d=1}^m R_d R_{0} ) + T R_{0} \sum_{i=1}^N n_i)$. Furthermore, the update of $\mathbf{U}_d$ requires $O(I_d \Pi_{d=1, d\neq k}^m I_d R_{task}\Pi_{d=1, d\neq k}^m R_d + I_d R_{task} \Pi_{d=1, d\neq k}^m R_d R_k  + T (\sum_{i=1}^N n_i) I_d R_d)$ steps with using a Lasso generalized linear model solver in Python (e.g., \texttt{sklearn.linear\_model.Lasso}). The update of $\mathcal{G}$ requires $O(\sum_{d=1}^m R_1 \ldots R_{k-1} I_k R_{k+1} R_m +R_{0} \Pi_{k=1}^m R_k + T (\sum_{i=1}^N n_i) R_{0} \Pi_{d=1}^m R_d)$. Moreover, the update of $\mathbf{D}_0$ requires $O(T (T_0 - Q_0) \sum_{i=1}^N n_i)$ operations, where each $\mathbf{d}_{0_i}$ is updated by solving a GLM with $(T_0 - Q_0)$ parameters for $n_i$ samples. The update of $\mathbf{V}_1$ requires $O(T T_1 p \sum_{i=1}^N n_i)$ operations using a lasso-regularized GLM solver, where $\mathbf{v}_1 \in \mathbb{R}^{T_1 p}$ is optimized over all $N$ individuals and their samples. Finally, the update of $\mathcal{H}$ requires $O(T T_0 T_1 \sum_{i=1}^N n_i)$ operations.

\subsection{Selection of Tuning Parameters}
\label{sec:tuning}

The proposed method involves selecting a few hyperparameters: $\mathbf{R} = [R_{0}, \ldots, R_{m}]$ (the decomposition rank for tensor predictors), $\mathbf{T} = [T_{0}, T_{1}]$ (the decomposition rank for vector predictors), and $\lambda$ (related to the lasso penalties for $\mathcal{G}$, $\mathcal{H}$, $\{\mathbf{U}_d\}$, $\mathbf{V}_1$). In particular, $R_{0}$ and $T_{0}$ are the ranks associated with the task subgroups, $R_d$ ($d =1, \ldots, m$) are the ranks for feature mode $d$ of tensor predictors, and $T_1$ is the rank for feature mode of scalar predictors. We perform a joint parameter selection mechanism to tune these hyperparameters. In particular, we employ five-fold cross-validation and select the ($\mathbf{R}$, $\mathbf{T}$, $\lambda$) tuple, which minimizes Root Mean Squared Error (RMSE). In the high-dimensional case, to simplify the computations, we let $\mathbf{R} = [R_{0}, R_{feature}, \ldots, R_{feature}]$. That is, $R_1 = \ldots = R_{m} = R_{feature}$.

\section{Performance Evaluation Using Simulation}
\label{sec:simulation}

This section evaluates the performance of the proposed method and provides a comprehensive comparison by using a simulated data set. We compare the proposed TenMTL method with (1) local models, where each task is learned individually based on the data available to that task, (2) A two-step approach, denoted by LR-Tucker, where the local models are first trained and the parameters are concatenated into a tensor. A Tucker decomposition is then applied to the concatenated model parameters to capture the similarity and correlation structure among the local model parameters, (3) global model, where a single global model is trained based on data available to all individuals, and (4) a multi-task learning method proposed in \cite{ando2005framework}, denoted by SVD-ASO. For local, global, and LR-Tucker benchmarks, a tensor regression framework is used. In the LR-Tucker approach, we smooth the trained local tensor regression models by applying a Tucker decomposition on the concatenated tensor of parameters. This is particularly helpful in situations where local model parameters are trained based on small sample sizes. Furthermore, SVD-ASO requires tuning a hyperparameter $\lambda$. Similar to TenMTL, we employ five-fold cross-validation to tune the $\lambda$.  We consider three simulation scenarios to evaluate the efficacy of the proposed method in comparison to the benchmarks. In the first scenario, we assess the performance of the methods when individuals are organized into distinct clusters. In the second scenario, we extend this setup to the case where individual models' correlations are reflected within specific groups of features rather than across all features. Finally, in the third scenario, we evaluate the performance of the methods in a high-dimensional data setting, where the inputs are higher-order tensors. 

\subsection{Data Generation}
This section describes the data generation procedure for the three simulation scenarios explained above. 

\noindent\textbf{Scenario I:}
This scenario simulates a setting where correlations between individual models extend across the entire feature set. That is, if two individual models (tasks) are similar, they exhibit similarity across all features. For each task $i$, we generate samples $\{y_{ij},\mathbf{x}_{ij}\}_{j=1}^{n_i}$, where $y_{ij} \in \mathbb{R}$ is the output and $\mathbf{x}_{ij} \in \mathbb{R}^{d}$ is the vector of covariates in the $j^{th}$ sample. Also, $d$ is the number of features, and $n_i$ is the sample size available to task $i$. 
The underlying data generation uses a linear model as, $y_{ij} = \beta_i^\top\mathbf{x}_{ij} + e_{ij}$, where $\beta_i$ is the underlying model parameter for individual $i$, and $e_{ij}$ is generated from a normal distribution with mean of zero and the variance of $\sigma_e^2$ to represent the noise. We assume individuals are grouped into $K$ clusters, where each cluster encompasses similar individuals. Particularly, individuals' data that belong to the same cluster follow the same distribution. To generate the data, each individual is randomly assigned to one of $K$ clusters with the probability of $\frac{1}{K}$. Let us denote by $c_i$ the cluster of task $i$. Next, we generate inputs $\mathbf{x}_{ij}$ from a normal distribution, $\mathbf{x}_{ij} \sim \mathcal{N}(h_{c_i}\mathbf{1}_d, \sigma^2_{x}\mathbf{I}_{d\times d})$, where $\mathbf{1}_d$ is a $d$-vector of ones and $\mathbf{I}_{d\times d}$ is an identity matrix. Here, $h_{c_i}$ is generated from a normal distribution $\mathcal{N}(1,\beta_x^2)$. 
To generate true underlying model parameters $\beta_i$ for each task, we first generate the tensor of $\mathcal{B}$ that contains all the model parameters. For this purpose, we generate a sparse core matrix $\mathcal{G}$, along with latent matrices $\mathbf{U}_{0} \in \mathbb{R}^{N\times R_0}$ and $\mathbf{U}_{1} \in \mathbb{R}^{d\times R_1}$. We generate the entries of the core tensor $\mathcal{G}=[g_{ij}] \in \mathbb{R}^{R_0 \times R_1}$ from a standard normal distribution, i.e., $g_{ij} \sim \mathcal{N}(0,1)$ with $s \%$ of its elements set to zero to enforce sparsity. Each row of $\mathbf{U}_{0}$, denoted as $ \mathbf{u}_{0_i} \in \mathbb{R}^{R_0}$, is drawn from a normal distribution with a cluster specific mean $p_{c_i}$ and standard deviation $\sigma_{u}$, i.e., $\mathbf{u}_{i_0} \sim \mathcal{N}(p_{c_i}\mathbf{1}_{R_0}, \sigma^2_{u}\mathbf{I}_{R_0\times R_0})$, where $p_{c_i} \sim \mathcal{N}(1, \beta_u^2)$. Here, $\beta_u$ controls the similarity between individual models, with higher values indicating greater heterogeneity. Next, the entries of $\mathbf{U}_{1}$ are simulated from a standard normal distribution. As the final step, $\mathcal{B}$ is constructed by $\mathcal{B}=\mathcal{G}\times_1\mathbf{U}_{0} \times_2 \mathbf{U}_{1}$. The $i^{th}$ row of $\mathcal{B}$ is the transpose of the model parameter for individual $i$. 

\noindent\textbf{Scenario II:} This scenario simulates a structured latent factor model where individual model correlations are reflected within specific groups of features rather than across all features. That is, if two individual models are similar, they share similarities in certain latent dimensions rather than uniformly across all features. We follow a similar strategy as in scenario 1 to generate input vectors $\mathbf{x}_{ij}$. Similar to the first scenario, to generate the model parameters, we construct $\mathcal{B}$ based on a sparse core matrix $\mathcal{G}$, and two latent factor matrices, $\mathbf{U}_{0}$ and $\mathbf{U}_{1}$, each exhibiting a \textit{column-wise group structure}. We generate the entries of $\mathcal{G}$ from a standard normal distribution, with $s \%$ of its elements replaced by zero to enforce sparsity. The factor matrices $\mathbf{U}_{0} $ and $\mathbf{U}_{1}$ are generated with column-wise group structures. Specifically, when generating $\mathbf{U}_0$, for each column $k=1,2,\ldots R_0$, a subset of tasks is randomly selected and assigned to a similarity group denoted by $\mathcal{I}_0^k\subseteq \{1,2,\ldots, N\}$. let $\mathbf{u}_{\mathcal{I}_0^k}$ denote the selected elements of the $k^{th}$ column of $\mathbf{U}_0$. Then, the elements of $\mathbf{u}_{\mathcal{I}_0^k}$ are generated from normal distribution $\mathcal{N}(\beta_{\text{group}}, \sigma^2_{\text{group}})$, while the entries that are not selected are drawn from a normal distribution with mean zero and a standard deviation $\sigma^2_{\text{non-group}}$. Similarly, when constructing $ \mathbf{U}_1$, for each column $k=1,\ldots, R_1$, we randomly select a subset of feature indices denoted by $\mathcal{I}_1^k$, and generate the entries of the column by $\mathcal{N}(\beta_{\text{group}}, \sigma^2_{\text{group}})$, if it is in $\mathcal{I}_1^k\subseteq \{1,2,\ldots, d\}$ and otherwise from a normal distribution with mean zero and a standard deviation $\sigma^2_{\text{non-group}}$. Here, the parameter $\beta_{\text{group}}$ controls intra-group similarity in the latent variables, which ensures that individual models assigned to the same group exhibit correlated latent factor values within specific feature groups rather than across all features. The final model parameter matrix is computed as: $\mathcal{B} = \mathcal{G}\times_1\mathbf{U}_0 \times_2 \mathbf{U}_{1}$. 

\noindent\textbf{Scenario III:} This scenario simulates a high-dimensional data setting where individual model correlations span the entire feature set. For each individual $i$, we generate a set of samples $\{y_{ij}, \mathcal{X}_{ij}\}_{j=1}^{n_i}$. Similar to the first scenario, we assume there are $K$ clusters that group similar individuals. Each individual is randomly assigned to a cluster with the probability of $\frac{1}{K}$. First the entries of an $m$-dimensional tensor input data $\mathcal{X}_{ij}$ is generated from a normal distribution with the mean value of $h_{c_i}$ and the variance of $\sigma^2_x$, where $h_{c_i} \sim \mathcal{N}(0, \beta_x^2)$. To generate $\mathcal{B}$, we first create a sparse core tensor $\mathcal{G} \in \mathbb{R}^{R_{0} \times R_1 \times \ldots \times R_{m}}$ where a specified proportion ($s\%$) of its entries are randomly set to zero, and the remaining entries are sampled from a standard normal distribution. For the individual-specific mode, the entries of a latent factor matrix $\mathbf{U}_{0} \in \mathbb{R}^{N \times R_{0}}$ are generated from a normal distribution, with the mean $p_{c_i}$ and variance $\sigma_u^2$, where $p_{c_i} \sim \mathcal{N}(1, \beta_u^2)$. For the remaining modes of the tensor, factor matrices $\mathbf{U}_{1}, \ldots, \mathbf{U}_{m}$ are generated with entries drawn from a standard normal distribution. The coefficient tensor $\mathcal{B}$ is then formed by mode-wise tensor contractions of the core tensor $\mathcal{C}$ with the factor matrices. That is, $\mathcal{B} = \mathcal{G} \times_1 \mathbf{U}_{0} \times_2 \mathbf{U}_1 \ldots \times_{m+1} \mathbf{U}_m$. 

\subsection{Simulation Specifications}
\label{sec:SimSpec}
In the first simulation scenario, we generate $N = 15$ tasks, each with 30 training and 30 test samples, and $d = 20$ features. The tasks are assigned to $K = 3$ clusters. We set the input parameters to $\beta_x = 0$, $\sigma_x = 0.1$, and $\sigma_u = 0.1$, and define the true rank as $R = [3, 4]$, where $R_0 = 3$ corresponds to the task mode and $R_1 = 4$ corresponds to the feature mode. We assess the model performance under two levels of observation noise $\sigma_e \in [0.5, 1]$ and three levels of sparsity $s \% \in [20, 40, 60]$. To evaluate the impact of heterogeneity across tasks, we vary $\beta_u \in [0, 0.2, 0.5, 1]$, where $\beta_u = 0$ indicates homogeneity (i.e., model parameters are drawn from the same distribution), and increasing values of $\beta_u$ introduce higher levels of heterogeneity. In the second scenario, we adopt a similar setting as Scenario I with $N = 15$ tasks, each having 30 training and 30 test samples with $d = 20$ features. We again use $K = 3$ clusters. However, here the signal is structured to follow a group-sparse pattern. We set $\beta_{group} = 1$, $\sigma_{group} = 0.1$, and select 8 feature groups. The model performance is evaluated for the same combinations of $\sigma_e \in [0.5, 1]$ and $ s \% \in [20, 40, 60]$. In the third simulation scenario, we consider tensor inputs, where each task has two modes of features with dimensions $d = [4, 5]$. A total of $N = 10$ tasks are generated, each with 30 training and 30 test samples. Tasks are drawn from $K = 2$ clusters, and the true rank is specified as $R = [2, 2, 2]$. We set $\beta_x = 0$, $\sigma_x = 0.1$, $\sigma_u = 0.1$, and $\sigma_e = 0.1$. To assess robustness against task heterogeneity and sparsity, we vary $\beta_u \in [0.2, 0.5, 1]$ and $s \% \in [20, 40, 60]$.

In all simulation studies, we set the maximum number of iterations to 50 and the convergence tolerance to $\epsilon = 10^{-5}$. We perform five-fold cross-validation to select the regularization parameter $\lambda$ from the candidate set $[0.0001, 0.0005, 0.001, 0.005, 0.01, 0.05, 0.1]$. For the first two simulation scenarios, the ranks $R_0$ and $R_1$ are selected from the candidate values $\{2, 3, 4, 5\}$ for each dimension. In the third simulation scenario, we choose the task mode rank $R_0$ and the feature mode rank $R_{\text{feature}}$ from the candidate values $\{2, 3, 4\}$, and set the full rank vector as $\mathbf{R} = [R_0, R_{\text{feature}}, R_{\text{feature}}]$. Model performance is evaluated using root mean squared error (RMSE). For all the experiments, we repeat the process thirty times to obtain the standard deviation of the RMSE.

\subsection{Simulation Results}
\label{sec:SimRes}

In this section, we apply the proposed method and the benchmarks to the simulated data and compare the performance of the methods based on the root mean squared error (RMSE). The results are provided for each data generation scenario as follows.
\paragraph{Scenario I:} Tables \ref{tab:sce1beta0}, \ref{tab:sce1beta0po2}, \ref{tab:sce1beta0po5}, and \ref{tab:sce1beta1} present the performance of the proposed method and the benchmarks in terms of RMSE taken over the test data of the $N=15$ individuals and over thirty replications, when the data is generated under Scenario I. The tables report the results for $\beta_u$ values of 0, 0.2, 0.5, and 1, respectively, and for different levels of sparsity ($s$) and noise ($\sigma_e$). In particular, $\beta_u=0$ corresponds to the case where the true individual task model parameters are drawn from the same distribution, and as $\beta_u$ increases, the heterogeneity among different task models increases. As it is reported in the tables, the proposed method outperforms all the benchmarks in almost all cases, particularly when $\beta_u$ takes a larger value. As an example, the mean (and standard deviation) RMSE values of TenMTL, LR-Tucker, Local, Global and SVD-ASO are 0.548 (0.083), 1.008 (0.566), 0.893 (0.203), 0.631 (0.166), and 3.557 (1.357), when $\beta_u=0$, $\sigma=0.5$, and $s = 0.4$. When the $\beta_u$ increases to one ($\beta_u=1$), for the same values of $\sigma=0.5$ and $s=0.4$, the RMSE values of TenMTL, LR-Tucker, Local, Global and SVD-ASO become 0.554 (0.090), 1.020 (0.722), 0.890 (0.204), 2.623 (1.739), and 4.438 (3.637). Note that the performance of the global model declines significantly with the increase in $\beta_u$. 

\begin{table}
\centering
\caption{Performance comparison of the methods in terms of RMSE when individual models are drawn from \underline{same distribution} ($\beta_u=0$)\label{tab:sce1beta0}}
\renewcommand{\arraystretch}{0.85}
\begin{tabular}{ccccccc}
\hline
$\sigma_e$& s &  TenMTL& LR-Tucker&  Local& Global & SVD-ASO           \\ \hline
0.5&0.2&  \textbf{0.559} (0.087) & 0.882 (0.462) & 0.896 (0.206) & 0.656 (0.180) & 3.853 (1.575)\\
          0.5&0.4& \textbf{0.548} (0.083) & 1.008 (0.566) & 0.893 (0.203) & 0.631 (0.166) & 3.557 (1.357) \\
          0.5&0.6&  \textbf{0.543} (0.080) & 1.058 (0.635) & 0.890 (0.209) & 0.588 (0.126) & 2.765 (1.329)\\
          1&0.2& \textbf{1.075} (0.151) & 1.771 (0.569) & 1.791 (0.413) & 1.103 (0.178) & 3.960 (1.543)\\
          1&0.4&  \textbf{1.074} (0.166) & 1.764 (0.556) & 1.789 (0.411) & 1.088 (0.169) & 3.673 (1.324) \\
        1& 0.6& \textbf{1.065} (0.156) & 1.657 (0.554) & 1.780 (0.407) & 1.058 (0.144) & 2.877 (1.208) \\ \hline
\end{tabular}
\end{table}

\begin{table}
\centering
\caption{Performance comparison of the methods in terms of RMSE when individual models have \underline{low} heterogeneity ($\beta_u=0.2$)\label{tab:sce1beta0po2}}
\renewcommand{\arraystretch}{0.85}
\begin{tabular}{ccccccc}
\hline
$\sigma_e$& s &  TenMTL& LR-Tucker&  Local& Global & SVD-ASO            \\ \hline 
0.5&0.2&  \textbf{0.556} (0.080) & 0.914 (0.579) & 0.895 (0.206) & 0.864 (0.343) & 3.968 (2.000)\\
          0.5&0.4& \textbf{0.551} (0.085) & 0.885 (0.461) & 0.893 (0.203) & 0.819 (0.325) & 3.635 (1.670) \\
          0.5&0.6&  \textbf{0.545} (0.082) & 1.101 (0.604) & 0.888 (0.201) & 0.713 (0.277) & 2.775 (1.436)\\
          1&0.2& \textbf{1.080} (0.153) & 1.802 (0.645) & 1.790 (0.414) & 1.250 (0.289) & 4.079 (1.960)\\
          1&0.4&  \textbf{1.062} (0.148) & 1.763 (0.552) & 1.789 (0.412) & 1.219 (0.274) & 3.754 (1.628) \\
        1& 0.6& \textbf{1.082} (0.168) & 1.568 (0.529) & 1.780 (0.408) & 1.145 (0.240) & 2.934 (1.371) \\ \hline
\end{tabular}
\end{table}

\begin{table}
\centering
\caption{Performance comparison of the methods in terms of RMSE when individuals have \underline{moderate} heterogeneity ($\beta_u=0.5$)\label{tab:sce1beta0po5}}
\renewcommand{\arraystretch}{0.85}
\begin{tabular}{ccccccc}
\hline
$\sigma_e$& s &  TenMTL& LR-Tucker&  Local& Global & SVD-ASO       \\ \hline
0.5&0.2&  \textbf{0.559} (0.081) & 0.890 (0.546) & 0.891 (0.206) & 1.536 (0.901) & 4.179 (2.928)\\
          0.5&0.4& \textbf{0.552} (0.091) & 1.026 (0.650) & 0.890 (0.202) & 1.434 (0.834) & 3.785 (2.444) \\
          0.5&0.6&  \textbf{0.546} (0.079) & 0.938 (0.484) & 0.884 (0.201) & 1.18 (0.738) & 2.889 (1.954)\\
          1&0.2& \textbf{1.084} (0.158) & 1.700 (0.622) & 1.788 (0.412) & 1.815 (0.808) & 4.313 (2.856)\\
          1&0.4&  \textbf{1.068} (0.152) & 1.709 (0.616) & 1.786 (0.412) & 1.728 (0.741) & 3.927 (2.370) \\
        1& 0.6& \textbf{1.086} (0.168) & 1.586 (0.574) & 1.778 (0.408) & 1.498 (0.647) & 3.069 (1.860) \\ \hline
\end{tabular}
\end{table}

\begin{table}
\centering
\caption{Performance comparison of the methods in terms of RMSE when individuals have \underline{high} heterogeneity ($\beta_u=1$)\label{tab:sce1beta1}}
\renewcommand{\arraystretch}{0.85}
\begin{tabular}{ccccccc}
\hline
$\sigma_e$& s &  TenMTL& LR-Tucker&  Local& Global & SVD-ASO           \\ \hline 
0.5&0.2&  \textbf{0.563} (0.086) & 0.888 (0.576) & 0.893 (0.208) & 2.828 (1.890) & 4.963 (4.352)\\
          0.5&0.4& \textbf{0.554} (0.090) & 1.020 (0.722) & 0.890 (0.204) & 2.623 (1.739) & 4.438 (3.637) \\
          0.5&0.6&  \textbf{0.548} (0.081) & 0.949 (0.521) & 0.884 (0.202) & 2.043 (1.554) & 3.411 (2.811)\\
          1&0.2& \textbf{1.088} (0.161) & 1.659 (0.683) & 1.790 (0.415) & 3.014 (1.802) & 5.099 (4.275)\\
          1&0.4&  \textbf{1.083} (0.164) & 1.692 (0.736) & 1.787 (0.414) & 2.821 (1.650) & 4.582 (3.558) \\
        1& 0.6& \textbf{1.074} (0.161) & 1.637 (0.637) & 1.780 (0.409) & 2.289 (1.457) & 3.589 (2.712) \\ \hline
\end{tabular}
\end{table}

\paragraph{Scenario II:} Table \ref{tab:sim2} reports the performance of each method in terms of average RMSE when data is generated under Scenario II. In this scenario, individual models have grouped correlation structures based on a group of features. As it is reported in this table, the proposed method (TenMTL) outperforms the benchmarks under all settings. For instance, the mean (and standard deviation) RMSE values of TenMTL, LR-Tucker, Local, Global, and SVD-ASO are 1.112 (0.174), 1.318 (0.389), 1.734 (0.419), 1.890 (0.849), and 2.268 (0.984). The superior performance of the proposed method lies in its ability to capture the cross-correlation structures and model the individual similarities that are based on a subset of features.

\begin{table}
\centering
\caption{Performance comparison of the methods in terms of RMSE when individuals have \underline{grouped correlation}\label{tab:sim2}}
\renewcommand{\arraystretch}{0.85}
\begin{tabular}{ccccccc}
\hline
$\sigma_e$& s &  TenMTL& LR-Tucker&  Local& Global & SVD-ASO          \\ \hline
0.5&0.2&  \textbf{0.561} (0.085) & 0.679 (0.152) & 0.868 (0.206) & 1.752 (0.906) & 2.196 (1.017)\\
          0.5&0.4& \textbf{0.565} (0.096) & 0.691 (0.187) & 0.866 (0.208) & 1.633 (0.907) & 2.069 (1.044) \\
          0.5&0.6&  \textbf{0.558} (0.088) & 0.717 (0.261) & 0.858 (0.204) & 1.390 (0.810) & 1.681 (0.892)\\
          1&0.2& \textbf{1.139} (0.185) & 1.326 (0.333) & 1.738 (0.418) & 1.992 (0.855) & 2.381 (0.964)\\
          1&0.4&  \textbf{1.112} (0.174) & 1.318 (0.389) & 1.734 (0.419) & 1.890 (0.849) & 2.268 (0.984) \\
        1& 0.6& \textbf{1.115} (0.169) & 1.333 (0.375) & 1.730 (0.418) & 1.683 (0.741) & 1.924 (0.816) \\ \hline
\end{tabular}
\end{table}

\paragraph{Scenario III:} Table \ref{tab:sim3} summarizes the performance of each method in terms of average RMSE when the data is generated under Scenario III. In this scenario, the input data for each task is high-dimensional. For example, the mean (and standard deviation) RMSE values of TenMTL, LR-Tucker, Local, Global, and SVD-ASO are 0.108 (0.015), 0.143 (0.049), 0.247 (0.099), 0.959 (0.979), and 3.023 (2.449). TenMTL outperforms the benchmark methods in a high-dimensional data setting.

\begin{table}
\centering
\caption{Performance comparison of the methods in terms of RMSE when individual models are \underline{higher-dimensional}\label{tab:sim3}}
\renewcommand{\arraystretch}{0.85}
\begin{tabular}{ccccccc}
\hline
$\beta_u$& s $\%$ &  TenMTL& LR-Tucker&  Local& Global & SVD-ASO            \\ \hline 
0.2&0.2& \textbf{0.110} (0.017) & 0.170 (0.109) & 0.253 (0.095) & 0.588 (0.474) & 3.424 (2.112) \\
          0.2&0.4& \textbf{0.109} (0.016) & 0.159 (0.076) & 0.252 (0.113) & 0.496 (0.438) & 3.047	(1.962)\\
          0.2&0.6& \textbf{0.108} (0.015) & 0.144 (0.055) & 0.230 (0.096) & 0.448 (0.434) & 2.709 (1.880) \\
          0.5&0.2& \textbf{0.109} (0.016) & 0.151 (0.096) & 0.251 (0.101) & 1.116 (1.035) & 3.389 (2.580) \\
          0.5&0.4&  \textbf{0.108} (0.015) & 0.143 (0.049) & 0.247 (0.099) & 0.959 (0.979) & 3.023 (2.449) \\
        0.5& 0.6&  \textbf{0.107} (0.015) & 0.143 (0.070) & 0.231 (0.098) & 0.876 (1.002) & 2.706 (2.309) \\ 
        1&0.2& \textbf{0.109} (0.017) & 0.162 (0.088) & 0.248 (0.099) & 2.075 (2.030) & 3.737 (3.509)\\
          1&0.4& \textbf{0.108} (0.015) & 0.154 (0.070) & 0.244 (0.104) & 1.804 (1.923) & 3.353 (3.418) \\
        1& 0.6& \textbf{0.109} (0.016) & 0.145 (0.058) & 0.225 (0.091) & 1.648 (1.981) & 3.014 (3.230) \\ \hline
\end{tabular}
\end{table}

\section{Case Study}
\label{sec:case}

This section introduces two case studies. The objective is to validate the performance of TenMTL with real data sets. In the first case study, we apply the proposed approach to a Parkinson's disease (PD) telemonitoring dataset to predict PD severity \citep{bot2016mpower}. In the second case study, we applied the proposed method to the attention deficit hyperactivity
disorder (ADHD) data from the ADHD-200 Sample Initiative \citep{bellec2017neuro}.

\subsection{Parkinson's Disease Severity Prediction Using mPower Telemonitoring Dataset}
\label{sec:mpower}

Parkinson’s disease (PD) is a complex neurological disorder marked by the degeneration of dopamine-producing cells in the midbrain. The symptoms include tremors, abnormal gait patterns, bradykinesia, and muscle stiffness. Affecting 7-10 million people globally, PD is the second most prevalent neurodegenerative condition after Alzheimer’s Disease \citep{pdnews2020}.  Although there is no cure for PD, effective treatment can significantly slow its progression, hence making regular monitoring and assessments critical for patients affected by PD. A widely used clinical tool for assessing the severity of PD is the Movement Disorder Society Unified Parkinson’s Disease Rating Scale (MDS-UPDRS) \citep{zhao2023robust}. This score is calculated based on a 65-question survey that is conducted at a specialized clinic. However, patients need to visit the clinic frequently to monitor the change in their PD condition, which may be challenging for most patients. Typically, patients visit the clinic every 4-6 months, and this is even more problematic for those living in distant areas with limited access to specialized care. Telemonitoring emerged as an effective healthcare tool to help with monitoring patients remotely. 
In this case study, we utilize a telemonitoring dataset gathered through the mPower application \citep{bot2016mpower}. The primary goal of this study is to develop a model that predicts MDS-UPDRS score based on activity data collected via smartphones. This model could then facilitate assessments of PD severity without requiring patients to visit those specialized clinics. 

One of the leading smartphone applications used for PD monitoring is mPower \citep{bot2016mpower}. Its popularity comes from the collection of patient data from extensive use of various sensors such as accelerometers, gyroscopes, and microphones. The application guides users through several pre-designed tasks, such as speaking and tapping. The tapping task evaluates participants' agility and pace by having them tap two fixed points on the screen with two fingers for 20 seconds. The voice activity involves recording participants' vocalization as they were instructed to repeatedly say 'Aaaaah' into the microphone. Then, several tapping and voice features are extracted from these recordings \citep{tsanas2011nonlinear, tsanas2012accurate, chaibub2016personalized}. Tapping features measure tapping speed, inter-tap interval, position, and features, while voice features measure amplitude, frequency, and noise. To effectively use the activity data gathered from the participants, a model is required to link the variations in activity data to the disease severity score. 

There are several challenges in using a telemonitoring dataset in predicting MDS-UPDRS for PD patients. First of all, the data collected by each participant has variations and some distinct patterns. This heterogeneity makes it challenging to build a global model by aggregating participants' data. Furthermore, each participant has a limited amount of data. Particularly, this activity data is high-dimensional with 339 voice features, 43 tapping features, and a sample size between 137 and 470 records collected over time. Thus, the local model will have limited generalization power and will be impaired by a lack of abundant samples. In this study, we consider each participant's predictive model as a task and utilize the capabilities of our proposed TensMTL approach in building a model when data is scarce and has heterogeneity.

\begin{figure}[H]
    \centering
    \caption{Performance comparison of the methods in predicting MDS-UPDRS}
    \includegraphics[trim=0.2cm 2cm 0.2cm 0.02cm,clip, width=3in]{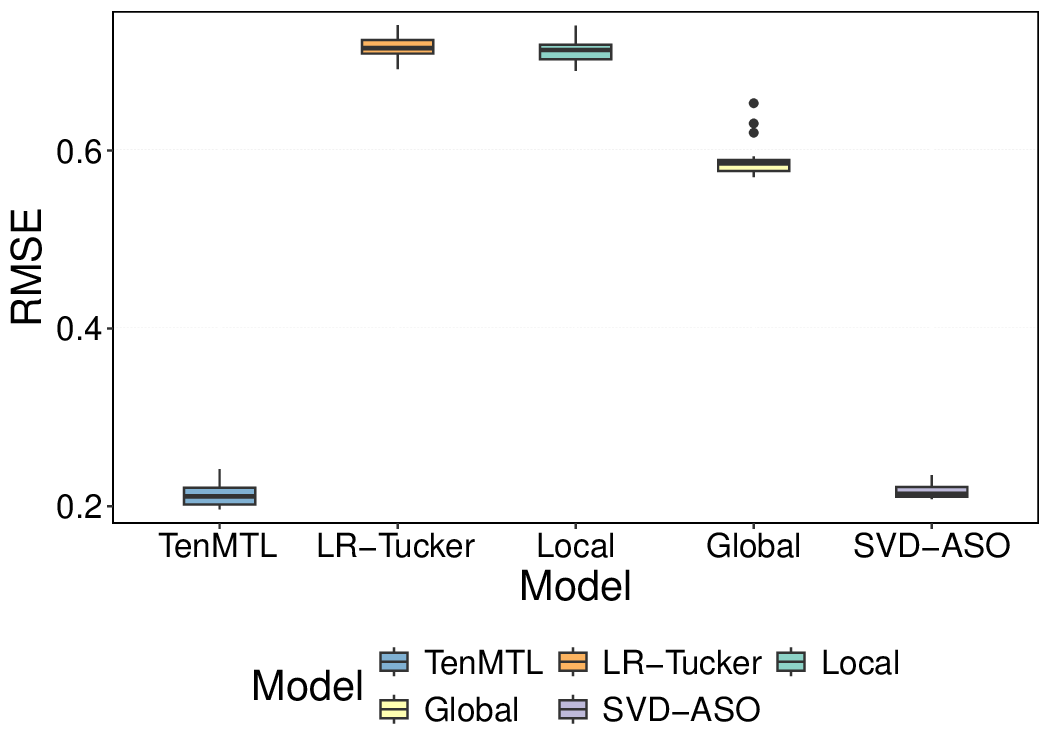}
    \label{fig:mpower}
\end{figure}

Figure \ref{fig:mpower} shows the box plots of the performance of the proposed methods and the benchmarks. Each boxplot shows the variability of the average RMSE (taken over $N$ individuals) across thirty replications. As shown in the boxplot, the proposed TenMTL model achieves the lowest RMSE among all methods in predicting MDS-UPDRS scores. In contrast, traditional global and local models yield higher errors, which indicate suboptimal performance in the presence of heterogeneous and limited data. Furthermore, the mean RMSE of the methods are 0.212, 0.716, 0.711, 0.614, and 0.217 for TenMTL, LR-Tucker, Local, Global, and SVD-ASO, respectively. Moreover, the proposed method effectively uncovers subpopulations of individuals who exhibit similar patterns in clinically relevant features, enabling a more comprehensive understanding of Parkinson’s disease progression. One such group is characterized by coherent trends across a set of key variables that are also highly predictive of MDS-UPDRS scores including tapping-related features such as $\textit{meanTapInter}$ (mean inter-tap interval) and $\textit{DFA}$ (detrended fluctuation analysis) \citep{arora2014high}, as well as voice-related features such as $\textit{mean\_MFCC\_1st}$, $\textit{mean\_MFCC\_3rd}$, $\textit{mea\_MFCC\_5th}$, and $\textit{mean\_MFCC\_7th}$, which represent different Mel frequency Cepstral coefficients (MFCCs) \citep{tsanas2011nonlinear}. These features capture distinct aspects of motor and vocal function, which offer a multi-dimensional representation of Parkinson's disease symptoms. Specifically, $\textit{meanTapInter}$ and $\textit{DFA}$ quantify tapping performance and temporal dynamics and reflect motor agility and rhythm irregularities, while the MFCC features characterize the shape of the speech spectrum and reveal articulatory and phonatory changes linked to the disease. The shared patterns in these features suggest a subgroup of individuals with similar sensorimotor deterioration, which aligns well with the latent task structure discovered by the model. This grouping illustrates how the proposed framework facilitates personalized modeling by capturing shared progression dynamics across individuals.

\subsection{Attention Deficit Hyperactivity Disorder Data Analysis}
\label{sec:adhd}

To evaluate the effectiveness of the proposed methods, we conducted a case study using the ADHD-200 dataset, a publicly available resource developed by the Neuro Bureau \citep{bellec2017neuro}. Attention-Deficit/Hyperactivity Disorder (ADHD) is one of the most common neurodevelopmental disorders, which affects approximately 5–10\% of children worldwide. Its core symptoms, including persistent inattention, hyperactivity, and impulsivity, can significantly impact daily functioning and social interactions. Neuroimaging has emerged as a powerful tool for investigating structural and functional brain abnormalities associated with ADHD. Several studies suggest that individuals with ADHD exhibit atypical brain development, particularly in regions responsible for executive control, attention modulation, and motor planning \citep{bellec2017neuro}. Understanding these patterns is crucial for early diagnosis, subtyping, and the development of personalized interventions. The ADHD-200 Sample Initiative was launched to accelerate research in this area by providing a large-scale, open-access neuroimaging dataset. It includes resting-state functional MRI (rs-fMRI), structural MRI (sMRI), and extensive phenotypic information from 896 individuals across multiple international sites. While this multi-site nature introduces heterogeneity in imaging protocols, demographics, and diagnostic practices, it also provides a more realistic and challenging testing for machine learning methods. The ability to develop models that generalize across such variations is critical for real-world deployment in clinical settings.

In this study, we focus on the structural MRI modality and use the preprocessed release provided by the Neuro Bureau. We adopt the FSL-based preprocessing pipeline as outlined in \cite{purnima_kumar_2024}, which includes steps such as reorientation, cropping, bias field correction, registration, skull stripping, and resampling all scans to a standardized resolution of 128 × 128 × 128 voxels, followed by pixel standardization. This ensures consistency across subjects and sites while preserving anatomical integrity. To facilitate personalized modeling and subgroup-specific insights, we define eight tasks based on age and gender. Age is divided into four developmental stages: [7.09–9.33), [9.33–11.43), [11.43–13.92), and [13.92–26.31]. Following \cite{li2018tucker}, we apply Tucker downsizing to reduce the spatial dimensionality of each scan to 12 × 12 × 12, which helps computationally efficient learning while retaining essential anatomical information.

\begin{figure}[H]
    \caption{Performance comparison of the methods in predicting patients with ADHD}
    \centering
    \includegraphics[trim=0.2cm 2cm 0.2cm 0.02cm,clip, width=3in]{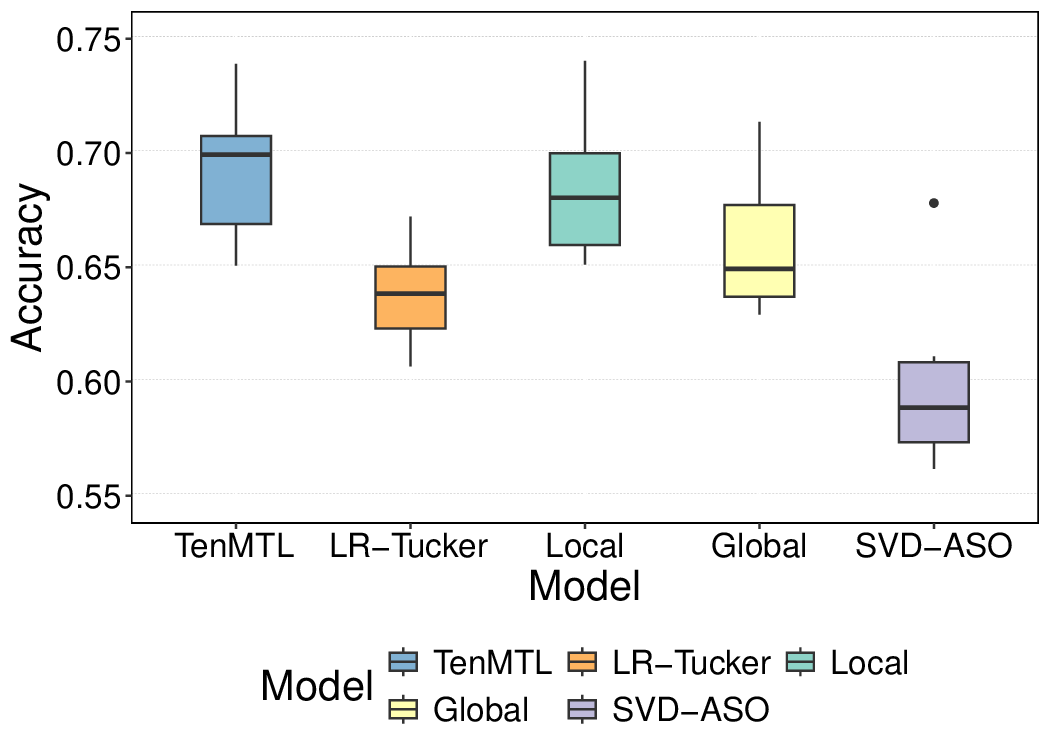}
    \label{fig:adhd}
\end{figure}

The use of a multi-task learning framework allows us to simultaneously model both shared and task-specific representations, and helps leverage common trends across subgroups while accommodating individual variation. Within each task, the data are randomly split into 75\% training and 25\% test sets. To evaluate predictive performance, we compare the proposed method against several baseline approaches across the eight defined tasks. Each method is assessed in terms of classification accuracy on the test sets, and the results are summarized in Figure \ref{fig:adhd}. These results underscore the advantage of incorporating subgroup structure and dimensionality reduction in modeling heterogeneous neuroimaging data for ADHD prediction. This case study not only demonstrates the effectiveness of our proposed methodology in a clinically relevant setting but also showcases its ability to scale to high-dimensional, heterogeneous neuroimaging data while preserving predictive performance.

\section{Conclusion}
\label{sec:conclusion}

This paper presents TenMTL, a novel framework for personalized modeling of heterogeneous subpopulations by integrating low-rank tensor decomposition within a multi-task learning (MTL) setting. In the proposed method, the collection of task-specific model parameters is represented as a higher-order tensor, which is then decomposed using Tucker decomposition. This formulation enables the joint modeling of both shared structures across tasks and individual-level variations, which allows for scalable and interpretable estimation of task-specific models. To learn the parameters, we define an objective function that balances model fit and low-rank regularization, and develop an alternating minimization algorithm that iteratively updates the core tensor and factor matrices. This approach not only reduces the number of effective parameters, as a result mitigating overfitting in high-dimensional settings, but also reveals latent components that capture commonalities and heterogeneity across tasks. 

Three simulation studies demonstrate the proposed method's superiority over existing benchmarks. Furthermore, we conduct two two case studies from the healthcare domain. In the first case study, we demonstrate the superior performance of TenMTL in predicting MDS-UPDRS scores of patients with Parkinson's Disease. In the second case study, we validate the practical effectiveness of TenMTL in classifying individuals with ADHD based on neuroimaging data. Beyond its predictive performance, TenMTL offers a structured way to separate shared and individual effects, which is important for analyzing high-dimensional data with limited samples, as often seen in fields like precision medicine and human-robot interaction. As future work, we plan to extend TenMTL to a federated learning setting, where tasks are distributed across decentralized nodes. This would allow the model to learn personalized structures without directly sharing sensitive data, making it well-suited for privacy-preserving applications.

\bibliography{ref}

%
\section{Appendix}

\subsection{Special Case: TenMTL for Vector Data}
\label{sec:appendix}

In this section, we introduce a special case of TenMTL, where the input data of the tasks is only in vector form. For instance, in telemonitoring applications such as the mPower dataset, each task may represent modeling a patient outcome, with input features derived from sensor-based walking or tapping activities and target outcomes reflecting clinical symptom scores \citep{bot2016mpower}. Suppose we have $N$ tasks corresponding to individuals or subgroups. Each task $i$ is associated with a set of $n_i$ scalar outcomes $y_{ij} \in \mathbb{R}$ and inputs (covariates) $\mathbf{x}_{ij} \in \mathbb{R}^d$, $(j=1, \ldots, n_i)$, where $n_i$ is the sample size for task $i$ and $d$ is the feature dimension. Similar to the general tensor case, we assume that $y_{ij} \vert \mathbf{x}_{ij}$ follows an exponential family distribution, $y_{ij} \vert \mathbf{x}_{ij} \sim f(y_{ij}; \theta_{ij})=\text{exp} \left( y_{ij}\theta_{ij} - b(\theta_{ij}) \right)$, where $b(.)$ is a distribution-specific known function and $\theta_{ij}$ is the canonical parameter. For the $i^{th}$ task, the goal is to learn a generalized linear model (GLM) to estimate the mean of the output with the following form $\theta_{ij}=\mu_i+\boldsymbol{\beta}_i^\top \mathbf{x}_{ij}$, where $\boldsymbol{\beta}_i \in \mathbb{R}^d$ is the model parameters, and $\mu_i$ is the intercept for task $i$. It is known that $b'(\theta_{ij})=E[y_{ij}\vert \mathbf{x}_{ij}]$. We simplify the notation by removing the intercept $\mu_i$. That is, we assume $\theta_{ij}=\boldsymbol{\beta}_i^\top \mathbf{x}_{ij}$. We construct tensor $\mathcal{B}=\begin{bmatrix}
    \boldsymbol{\beta}_1, \boldsymbol{\beta}_2, \ldots, \boldsymbol{\beta}_N
\end{bmatrix}^\top \in \mathbb{R}^{N \times d}$ and impose low-rankness to capture the similarity among tasks.  Suppose that $\mathcal{B}$ admits a Tucker decomposition of the form $\mathcal{B}=\mathcal{G} \times_1 \mathbf{U}_0 \times_2 \mathbf{U}_1$, where $\mathbf{U}_0 \in \mathbb{R}^{N \times R_0}$ is a factor matrix capturing correlations in models associated with each subgroup, $\mathbf{U}_1 \in \mathbb{R}^{d \times R_1}$ is the second factor matrix capturing correlations in the input variables (i.e., features), and $\mathcal{G} \in \mathbb{R}^{R_0 \times R_1}$ is the core tensor which captures the relationships between the subgroups and the features. We can matricize the Tucker decomposition form through the first mode as $ \mathcal{B}_{(0)}=\mathbf{U}_0 \mathcal{G}_{(0)}  \mathbf{U}_{1}^\top$, where $\mathcal{B}_{(0)}$ and $\mathcal{G}_{(0)}$ are the first mode matricizations of $\mathcal{B}$ and $\mathcal{G}$, respectively. With these assumptions and given training data, we estimate the model parameters by solving the following minimization problem:
\begin{equation}
    \begin{aligned}
        &\min_{\theta_{ij} } \; \sum_{i=1}^N \sum_{j=1}^{n_i} ( -y_{ij}\theta_{ij} + b(\theta_{ij}) ) \\
        &\;\; \emph{s.t.} \;\quad \theta_{ij}=\boldsymbol{\beta}_i^\top \mathbf{x}_{ij}, \forall i \\
        &\qquad\quad\; \mathcal{B}_{(0)}=\mathbf{U}_0\mathcal{G}_{(0)}  \mathbf{U}_{1}^\top \\
        &\qquad\quad\; \mathcal{B}_{(0)}=\begin{bmatrix}
    \boldsymbol{\beta}_1, \boldsymbol{\beta}_2, \ldots, \boldsymbol{\beta}_N
\end{bmatrix}^\top,
    \end{aligned}
\end{equation}
which is equivalent to solving the following problem:
\begin{equation}
\label{eq:opt}
        \min_{\{\mathbf{u}_{0_i}\},\mathbf{G}_{(0)},\mathbf{U}_{1}  } \; \sum_{i=1}^N \sum_{j=1}^{n_i} ( -y_{ij}\mathbf{u}_{0_i} \mathcal{G}_{(0)}  \mathbf{U}_{1}^\top \mathbf{x}_{ij} + b(\mathbf{u}_{0_i} \mathcal{G}_{(0)}  \mathbf{U}_{1}^\top \mathbf{x}_{ij}) ),
\end{equation}
where $\mathbf{u}_{0_i} \in \mathbb{R}^{1 \times R_0} $ is the $i^{th}$ row of $\mathbf{U}_{0}$. To handle the subgroup-specific correlation and to have numerical stability, we add a lasso penalty to $\mathbf{G}_{(0)}$ and $\mathbf{U}_{1}$ as follows: 
\begin{equation}
\label{eq:opt_ridge}
\begin{aligned}
    \min_{\{\mathbf{u}_{0_i}\},\mathbf{G}_{(0)},\mathbf{U}_{1}  } \; &\sum_{i=1}^N \sum_{j=1}^{n_i} ( -y_{ij}\mathbf{u}_{0_i} \mathcal{G}_{(0)}  \mathbf{U}_{1}^\top \mathbf{x}_{ij} + b(\mathbf{u}_{0_i} \mathcal{G}_{(0)}  \mathbf{U}_{1}^\top \mathbf{x}_{ij}) ) + \lambda_g \vert\vert \mathcal{G}_{(0)} \vert\vert_1 + \lambda_u \vert\vert \mathbf{U}_{1} \vert\vert_1.
\end{aligned}  
\end{equation}

To solve problem \eqref{eq:opt_ridge}, we employ the alternating least squares (ALS) algorithm that iteratively updates the variables as follows: 
\noindent\textbf{\\Update of \texorpdfstring{$\mathbf{u}_{0_i}$}{Lg}: } When updating $\mathbf{u}_{0_i}$, the problem reduces to: 
\begin{equation}
\label{eq:updateu1}
        \min_{\mathbf{u}_{0_i} } \; l = \sum_{j=1}^{n_i} ( -y_{ij}\mathbf{u}_{0_i} \mathbf{k}_{ij} + b(\mathbf{u}_{0_i} \mathbf{k}_{ij}) ),
\end{equation}
where $\mathbf{k}_{ij}:=\mathcal{G}_{(0)}\mathbf{U}_{1}^\top\mathbf{x}_{ij} \in \mathbb{R}^{R_1 \times 1}$. We solve the problem for $\mathbf{u}_{0_i}$, $\forall i$, with generalized linear model (GLM) solver.
\noindent\textbf{\\Update of \texorpdfstring{$\mathbf{U}_{1}$}{Lg}:} When updating $\mathbf{U}_{1}$, the problem reduces to:
\begin{equation}
        \min_{\mathbf{U}_{1}  } \; \sum_{i=1}^N \sum_{j=1}^{n_i} ( -y_{ij}\mathbf{t}_i  \mathbf{U}_{1}^\top \mathbf{x}_{ij} + b(\mathbf{t}_i  \mathbf{U}_{1}^\top \mathbf{x}_{ij}) ) +\lambda_u \vert\vert  \mathbf{U}_{1} \vert\vert_1,
\end{equation}
where $\mathbf{t}_i:=\mathbf{u}_{0_i} \mathcal{G}_{(0)} \in \mathbb{R}^{1 \times R_1}$. By setting $\mathbf{u}_1:=\text{vec}(\mathbf{U}_{1}^\top) \in \mathbb{R}^{R_1d}$ and $\mathbf{h}_{ij}:=\mathbf{x}_{ij}^\top \otimes \mathbf{t}_i \in \mathbb{R}^{1 \times R_1d}$, we solve the following lasso regularized generalized linear model (GLM):
\begin{equation}
\label{eq:updateU2}
        \min_{\mathbf{u}_1  } \; \sum_{i=1}^N \sum_{j=1}^{n_i} \left( -y_{ij}\mathbf{h}_{ij}\mathbf{u}_1 + b(\mathbf{h}_{ij}\mathbf{u}_1) \right) +\lambda_u \vert\vert \mathbf{u}_1 \vert\vert_1.
\end{equation}
\noindent\textbf{\\Update of \texorpdfstring{$\mathcal{G}_{(0)}$}{Lg}:} To update $\mathcal{G}_{(0)}$, we solve:
\begin{equation}
\begin{aligned}
     \min_{\mathcal{G}_{(0)} } \; &\sum_{i=1}^N \sum_{j=1}^{n_i} ( -y_{ij}\mathbf{u}_{0_i} \mathcal{G}_{(0)}  \mathbf{m}_{ij} + b(\mathbf{u}_{0_i} \mathcal{G}_{(0)}  \mathbf{m}_{ij} )) +\lambda_g \vert\vert \mathcal{G}_{(0)} \vert\vert_1,
\end{aligned}
\end{equation}
where $\mathbf{m}_{ij}:=\mathbf{U}_1^\top\mathbf{x}_{ij} \in \mathbb{R}^{R_1 \times 1}$. We then transform and solve the following lasso regularized GLM:
\begin{equation}
\label{eq:updateC}
        \min_{\mathbf{g}} \; \sum_{i=1}^N \sum_{j=1}^{n_i} ( -y_{ij}\mathbf{o}_{ij} \mathbf{g} + b(\mathbf{o}_{ij} \mathbf{g} ) ) +\lambda_g \vert\vert \mathbf{g} \vert\vert_1,
\end{equation}
where $\mathbf{g}:=\text{vec}(\mathcal{G}_{(0)}) \in \mathbb{R}^{R_0R_1}$, and $\mathbf{o}_{ij}:=\mathbf{m}_{ij}^\top \otimes \mathbf{u}_{0_i} \in \mathbb{R}^{1 \times R_0R_1}$.

This procedure is repeated until convergence, i.e., $\frac{l^t - l^{t-1}}{l^{t-1}} < \epsilon$ where $l^t$ is the objective function value at iteration $t$. After estimating $\mathbf{u}_{0_i}, ( i = 1, \ldots, N )$, $\mathcal{G}_{(0)}$, and $\mathbf{U}_{1}$, the parameters for individual models can be obtained by $\boldsymbol{\beta}_i^\top = \mathbf{u}_{0_i} \mathcal{G}_{(0)}  \mathbf{U}_{1}^\top$.

\end{document}